\documentclass[letterpaper,10pt,journal,twoside]{IEEEtran}

\IEEEoverridecommandlockouts                              

\usepackage{graphics} 
\usepackage{graphicx} 
\usepackage{amssymb}  
\usepackage{graphicx}
\usepackage{subfig} 
\usepackage{url}
\usepackage{float}
\usepackage{caption}
\usepackage{amsmath} 

\usepackage[table]{xcolor}
\usepackage{array}
\usepackage{hyperref}

\title{ \LARGE \bf Chalito: An Extensible Library for Filtering-Based State Estimation in Quadruped Robots }

\author{ Hilton Marques Souza Santana$^{1}$, João Carlos Virgolino Soares$^{2}$, Marco Antonio Meggiolaro$^{1}$ and \\Claudio Semini$^{2}$
\thanks{This work was partially financed by the Brazilian Federal Agency for Support and Evaluation of Graduate Education (CAPES), the Carlos Chagas Filho Foundation for Research Support of the State of Rio de Janeiro (FAPERJ).} 
\thanks{$^{1}$H. M. S. Santana and M. A. Meggiolaro are with the Department of Mechanical Engineering at the Pontifical Catholic University of Rio de Janeiro,
       {\tt\small hiltonmarquess@gmail.com} /
       {\tt\small meggi@puc-rio.br}}%
\thanks{$^{2}$ J. C. V. Soares and C. Semini are with the Dynamic Legged Systems Lab, Istituto Italiano di Tecnologia, {\tt\small joao.virgolino@iit.it} / 
{\tt\small claudio.semini@iit.it}}}

\begin{document}




\maketitle

\begin{abstract}

State estimation is essential for quadruped robots, enabling robust locomotion, navigation, and control. While many estimators have been proposed in the literature, existing implementations are often tied to specific robots or software stacks, making fair comparisons difficult. This lack of a general-purpose benchmarking framework hinders reproducibility and slows down algorithmic innovation. In this paper, we introduce Chalito, an extensible MATLAB/Python library for benchmarking filter-based state estimation algorithms in quadruped robots. Chalito imports robot models directly from URDF, supports multiple filtering approaches, and is designed to be easily extended with new methods. The framework runs on both simulated and real datasets, enabling systematic evaluation across robots and filters. To the best of our knowledge, this is the first open-source library exclusively dedicated to benchmarking filtering algorithms for quadruped robots.

\end{abstract}

\section{INTRODUCTION}

State estimation is critical for quadruped robots, as it is an essential step toward achieving robust control and enabling autonomous navigation~\cite{Hartley2019ContactaidedIE}. Accurate estimation of the robot’s pose, velocity, and contact states is fundamental for locomotion over unstructured or slippery terrain, interaction with the environment, and integration with higher-level planning and control.


To date, several classes of state-estimation filters have been applied to quadruped robots, and an increasing number of publicly available datasets have been released for different robotic platforms and operating scenarios. The most commonly adopted approaches include the Extended Kalman Filter (EKF)~\cite{camurri2020frontiers, nistico2025}, the Observability-Constrained EKF~\cite{bloesch2012}, the Unscented Kalman Filter~\cite{bloesch2013}, and the Invariant EKF~\cite{Hartley2019ContactaidedIE, lin2023}. 
In parallel, the growing availability of public datasets has enabled the evaluation of these methods across a variety of quadruped robots and environments. 
We show in Table~\ref{tab:quadruped_datasets} some sequences of public datasets known to the authors. Despite the progress achieved by these filtering approaches, there is currently no open-source framework that enables a systematic comparison of filter performance in terms of both accuracy and consistency. Likewise, there is no common benchmarking platform that allows researchers to easily evaluate and compare state-estimation algorithms across multiple quadruped robots and datasets. Existing implementations are often tightly coupled to specific robots, sensor suites, or software stacks (typically C++/ROS)~\cite{nistico2026}, making it difficult to conduct fair, systematic comparisons across algorithms, platforms, and datasets. This gap hinders reproducibility, slows down the evaluation of new algorithms, and limits accessibility for education and rapid prototyping.


In this work, we aim to address this gap by presenting a MATLAB/Python library for quadruped state estimation that is both URDF\footnote{URDF stands for Unified Robot Description format, which is a file that describes the robot's kinematic tree}-driven and filter-agnostic. Our framework enables researchers to load any quadruped robot model, select among a variety of existing filtering approaches, or include their own method in the library, and benchmark their performance. Being aware that MATLAB is not free software, 
we are also providing a Python implementation of the proposed library, called PyChalito. The main difference between Chalito and PyChalito is 
that while the former is focused on reading a fixed dataset (as described in Section \ref{sec:config_file}), the latter is tied to work along with MuJoCo~\cite{todorov2012} 
using the QuadrupedPyMPC library~\cite{turrisi2024}.

\begin{figure}[t!]
\centering
\includegraphics[width=0.48\textwidth]{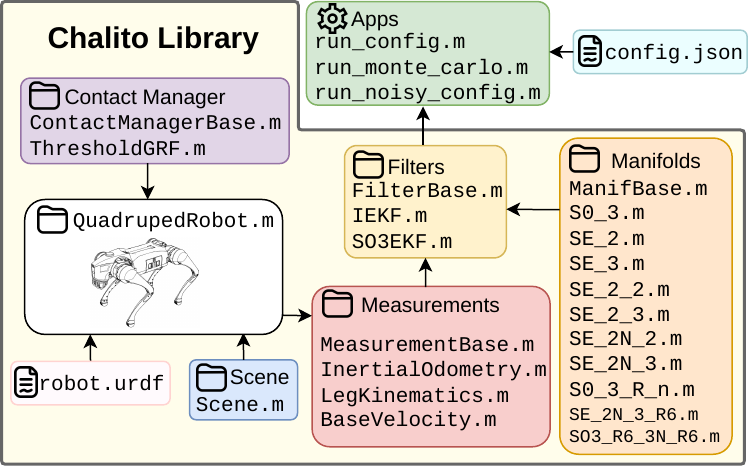}
\caption{MATLAB implementation of the Chalito architecture.}
\label{fig:main_modules}
\end{figure}

The main contributions of this paper are:

\begin{itemize}
    \item To the best of our knowledge, this is the first extensible open-source library focused on benchmarking quadruped robot filter-based state estimators in terms of accuracy and consistency. The provided interface architecture allows one to switch between different filters, robots, and contact managers easily.
    \item We describe in detail the main interfaces that compose the proposed library. These interfaces are the core for the extensibility property of the library.
      Moreover, MATLAB and Python implementations of these interfaces are also provided.
    \item We demonstrate the advantages of the proposed library through three evaluations, containing two filters and six different robots. 
      The first is a convergence analysis performed with the library running 
      on a MuJoCo simulation. The second is a Monte Carlo analysis performed with synthetic data. Finally, in the third evaluation, we use the library 
      to estimate the robot position in all distinct public real-world datasets described in Table \ref{tab:quadruped_datasets}.
\end{itemize}

The remainder of this paper is organized as follows: Section \ref{sec:literature_review} presents a review of the related literature, while 
Section \ref{sec:methodology} describes the three main interfaces and classes that compose the library: Lie group, Filter, Measurement Interfaces, and the Quadruped Robot class, also shown in 
Fig. \ref{fig:main_modules}.
Finally, Section \ref{sec:results} details three important applications of the library.

\section{Literature Review}
\label{sec:literature_review}
Filtering algorithms for quadruped robots are composed of the following three main parts: the non-linear filter, which encompasses the state and the manifold it resides on; the contact manager, which determines whether the robot’s feet are in contact with the ground; and the measurement module, which processes the synchronized measurements available within the framework. One of the main goals of Chalito is to abstract and expose the relevant interfaces of these three components, enabling users to more easily test how a new implementation of one part interacts with the others. The remainder of this section describes the related work of these three components, as well as related open-source libraries.

Over the past decade, numerous non-linear Kalman filtering methods have been proposed. While these methods differ in their formulations, many aim to represent the state in spaces that respect the system’s intrinsic geometric structure, which in some cases can be modeled as a non-linear manifold. Such representations help to enforce geometric constraints and promote consistent state and error propagation. One of the main breakthroughs in this direction was achieved by \cite{barrau2015}, who showed that the false observability problem in the EKF, 
first identified in \cite{huang2008}, is completely coordinate-dependent, and modifying the error parameterization resolves the issue.

There are two main C++ open-source libraries specialized in non-linear Kalman Filter based on manifolds: \texttt{IKFoM} \cite{he2021} and \texttt{manif} \cite{deray2020}. The first focuses on smooth manifolds and their Cartesian products, whereas the second is specialized in manifolds that are also matrix Lie groups. However, \texttt{IKFoM} does not handle semi-direct products of manifolds, which are required to represent the important family of Special Euclidean groups, and \texttt{manif} still lacks support for the Multiple Direct Spatial Isometries Group, $\mathrm{SE}_{2+N}(3)$, required for leg odometry \cite{Hartley2019ContactaidedIE}. Moreover, when deploying a filter for quadruped robots \cite{Hartley2019ContactaidedIE, camurri2020frontiers, nistico2025, lin2023, yang2022}, researchers often need to implement their own versions of these manifolds, making meaningful comparisons difficult. This challenge usually arises from the complexity of integrating with two other key modules: the contact manager and synchronized measurements. The proposed library was designed to address these gaps: it is general enough to encompass all relevant manifolds while remaining easy to use in legged-robot applications.

The contact manager is an active research area in its own right. Existing methods include logistic regression on ground reaction forces (GRF) \cite{camurri2017} and, more recently, inference using neural networks \cite{lin2023,lin2021}. In this work, we implemented the simplest method for estimating contact (GRF threshold) while abstracting the main interface of a contact manager, thereby enabling future testing of new methods without modifying the exposed interface.

A similar strategy was employed in the measurement module, where we proposed a unified interface to encompass all measurement modalities. Within this framework, the main measurements already established in the context of quadruped robots were consolidated into a single interface. These include IMU integration, leg odometry~\cite{Hartley2019ContactaidedIE, bloesch2012} and estimation of base velocity during contact~\cite{bloesch2013}. This design also opens the door to testing new types of measurements in the future.

Recently, \cite{nistico2026} presented a benchmark comparing three
state-of-the-art, real-time state estimators for quadruped robots.
However, because each estimator is provided as a standalone implementation,
 the framework is not designed to be extended. Adding a new
method requires reimplementing functionality common to all filters
rather than reusing it. Chalito targets a different setting. It is not
aimed at real-time deployment; instead, it is built on three pillars.
The first is code reuse: a unified interface for Lie groups,
robot-related operations, and measurements lets the user focus on the
specific novelty of a new filter instead of reimplementing
well-established components. Second is the clarity of matrix
operations, with an implementation kept as close as possible to the
mathematical formulation. The third is visualization: we provide three
representations of the robot, ranging from simplistic to realistic
(see Figs.~\ref{fig:robot_debug} and \ref{fig:exp_position}), to
facilitate debugging of the entire pipeline. Our guiding philosophy is
that a researcher can quickly evaluate a new method against previous
ones in Chalito, obtain a reference for the expected outputs, and then
deploy an optimized version on the robot.

\begin{table}[t]
\centering
\begin{tabular}{p{2cm} p{2cm} p{3cm}}
\hline
\textbf{Work} & \textbf{Robot} & \textbf{Dataset} \\
\hline

Pronto \cite{camurri2017} &
ANYmal B &
\href{https://drive.google.com/file/d/1a_BA7yyj4XdUcCXrxpz5o1PdCi0fJn5K/view}{Fire Service College}
\\

Cerberus 1.0 \cite{yang2022} &
Unitree A1 &
\href{https://drive.google.com/drive/folders/1PdCLMVRqS97Tc9VbJY12EUl9UHpVoO_X}{Street}
\\

Cerberus 2.0 \cite{yang2023} &
Unitree Go1 &
\href{https://drive.google.com/file/d/1THkuvGwgMDXAfQdN2d3mW3Xyg7KyDnSX/view}{Square}
\\

Leg-KILO \cite{legkilo} &
Unitree Go1 &
\href{https://drive.google.com/file/d/1QWehixg4zeefXr06x4ltVp7IzBusogtH/view?usp=drive_link}{Running}
\\

GrandTour \cite{grantour} &
ANYmal D &
\href{https://grand-tour.leggedrobotics.com/dataset\#mission-19-on}{Pilatus-Hike2}
\\

\hline
\end{tabular}
\caption{Publicly available quadruped robot dataset sequences tested in this work.}
\label{tab:quadruped_datasets}
\vspace{-4mm}
\end{table}

\section{METHODOLOGY} 
\label{sec:methodology}
\subsection{Lie group interface}
A Lie group $G$ is simultaneously a smooth manifold and a continuous group. These structures play a fundamental role in robotics, as they provide the natural extension of vector spaces to nonlinear and, in general, non-commutative manifolds. Much of the beauty of a Lie group relies on its Lie algebra, $\mathfrak{g}$, which is a vector space that encodes much of the structure of the Lie group (see Subsection \ref{sec:filter_interface}).   
In this work, we provide support for the family of matrix Lie groups listed in Table \ref{list_manifolds}. 
Recall that every matrix Lie group can be embedded as a closed subgroup of the General Linear Group, $\mathrm{Gl}(d, \mathbb{R})$, $d \in  \mathbb{N}$.
In Table \ref{list_manifolds} we show the dimension $n$ of each considered Lie group along with its common application in state estimation theory.
From an implementation standpoint, all manifolds are derived from a common base class, \path{ManifBase}, which defines the core functionalities and ensures a consistent interface across group representations. 

The primary inspiration for this class was the open-source C++ library \path{manif} \cite{deray2020}. Our aim was to preserve as much as possible of its nomenclature and structural design, thereby providing a natural extension to MATLAB/Python. A distinction, however, is that unlike \path{manif}, our implementation supports dynamic groups, such as $\mathrm{SE}_{2+N}(3)$, where $N$ is a runtime variable representing, for example, the number of feet in contact of a legged robot and/or the number of visible landmarks. Since \path{manif} is a template-based, compiled library, the size of its derived classes is fixed at compile time. 
A new design pattern that might solve this problem is currently being discussed in a \path{manif} GitHub issue\footnote{\url{https://github.com/artivis/manif/issues/328\#issuecomment-2760713471}}.
In contrast, our design enables such flexibility without sacrificing consistency with the established \path{manif} conventions.
\begin{table}[h!]
\centering
\rowcolors{2}{gray!15}{white}
\scalebox{0.91}{
\begin{tabular}{|c|c|c|}
\hline
\textbf{Matrix Lie groups} & $n$ & \textbf{Application} \\
\hline
$\mathrm{SO}(3)$ & $3$ &  Rotation (3D) \\
$\mathrm{SE}(2)$ & $4$ &  Rotation + Position (RP) (2D) \\
$\mathrm{SE}(3)$ & $6$ &  RP (3D) \\
$\mathrm{SE}_2(2)$ & $6$ &  RP + Velocity (RPV) (2D) \\
$\mathrm{SE}_2(3)$ & $9$ &  RPV (3D) \\
$\mathrm{SE}_{2+N}(2)$ & $6+2N$ & RPV + $N$ (FP or LM) (2D) \\
$\mathrm{SE}_{2+N}(3)$ & $9+3N$ & RPV + $N$ (FP or LM) (3D) \\
$\mathrm{SO}(3)\!\times\!\mathbb{R}^{6+3N}$ & $9+3N$ & RPV + $N$ (FP or LM) (3D) \\
$\mathrm{SE}_{2+N}(3) \times \mathbb{R}^{6} $ & $9+3N + 6$ & RPV + $N$ (FP or LM) + IMU Biases \\
$\mathrm{SO}(3)\!\times\!\mathbb{R}^{12+3N}$ & $15+3N$ & RPV + $N$ (FP or LM) + IMU Biases \\
\hline
\end{tabular}
}
\caption{List of the main Matrix Lie groups implemented in Chalito and their associated 
state estimation application. FP stands for Foot Position, while LM stands for Landmark.}
\label{list_manifolds}
\end{table}
The \texttt{ManifBase} class provides the interface for the main operations on Lie groups (see Table \ref{lie_group_operations}). A detailed description of each operation can be found in \cite{sola2018}.
\begin{table}[ht]
\centering
\rowcolors{2}{gray!15}{white}
\scalebox{0.9}{
\begin{tabular}{|c|c|}
\hline
\textbf{Operation} & \textbf{Interface's Code} \\
\hline
$G \to G,\ \mathcal{X} \mapsto \mathcal{X}^{-1}$ & \texttt{X.inverse()} \\
$G \times G \to G,\ (\mathcal{X},\mathcal{Y}) \mapsto \mathcal{X}\circ\mathcal{Y}$ & \texttt{X * Y} \\
$\mathbb{R}^n \to \mathfrak{g},\ \boldsymbol{\omega} \mapsto \boldsymbol{\omega}^\wedge$ & \texttt{G.hat(w)} \\
$\mathfrak{g} \to \mathbb{R}^n,\ \boldsymbol{\omega}^\wedge \mapsto (\boldsymbol{\omega}^\wedge)^\vee$ & \texttt{G.vee(W)} \\
$G \times \mathbb{R}^m \to \mathbb{R}^m,\ (\mathcal{X},\mathbf{v}) \mapsto \mathcal{X}\circ\mathbf{v}$ & \texttt{X.act(v)} \\
$\mathbb{R}^n \to G,\ \boldsymbol{\omega} \mapsto \exp(\boldsymbol{\omega}^\wedge)$ & \texttt{G.exp(w)} \\
$G \to \mathbb{R}^n,\ \mathcal{X} \mapsto \log(\mathcal{X})^\vee$ & \texttt{X.log()} \\
$G \to \mathbb{R}^{n\times n},\ \mathcal{X} \mapsto \mathrm{Adj}(\mathcal{X})$ & \texttt{X.adj()} \\
$G \times \mathbb{R}^n \to G,\ (\mathcal{X},\boldsymbol{\omega}) \mapsto \mathcal{X} \oplus \boldsymbol{\omega}$ & \texttt{X + w} \\
$G \times \mathbb{R}^n \to G,\ (\mathcal{X},\boldsymbol{\omega}) \mapsto \boldsymbol{\omega} \oplus \mathcal{X}$ & \texttt{w + X} \\
$G \times G \to \mathbb{R}^n,\ (\mathcal{X},\mathcal{Y}) \mapsto \mathcal{X} \ominus \mathcal{Y}$ & \texttt{X - Y} \\
\hline
\end{tabular}
}
\caption{Main operations in the \texttt{ManifBase} class.}
\label{lie_group_operations}
\vspace{-6mm}
\end{table}
\subsection{Quadruped Robot class}
\begin{figure}[t!]
\centering
\subfloat[\label{fig:subdivision}]{
    \includegraphics[
        scale=0.1,
        trim=480 180 0 0, 
        clip
    ]{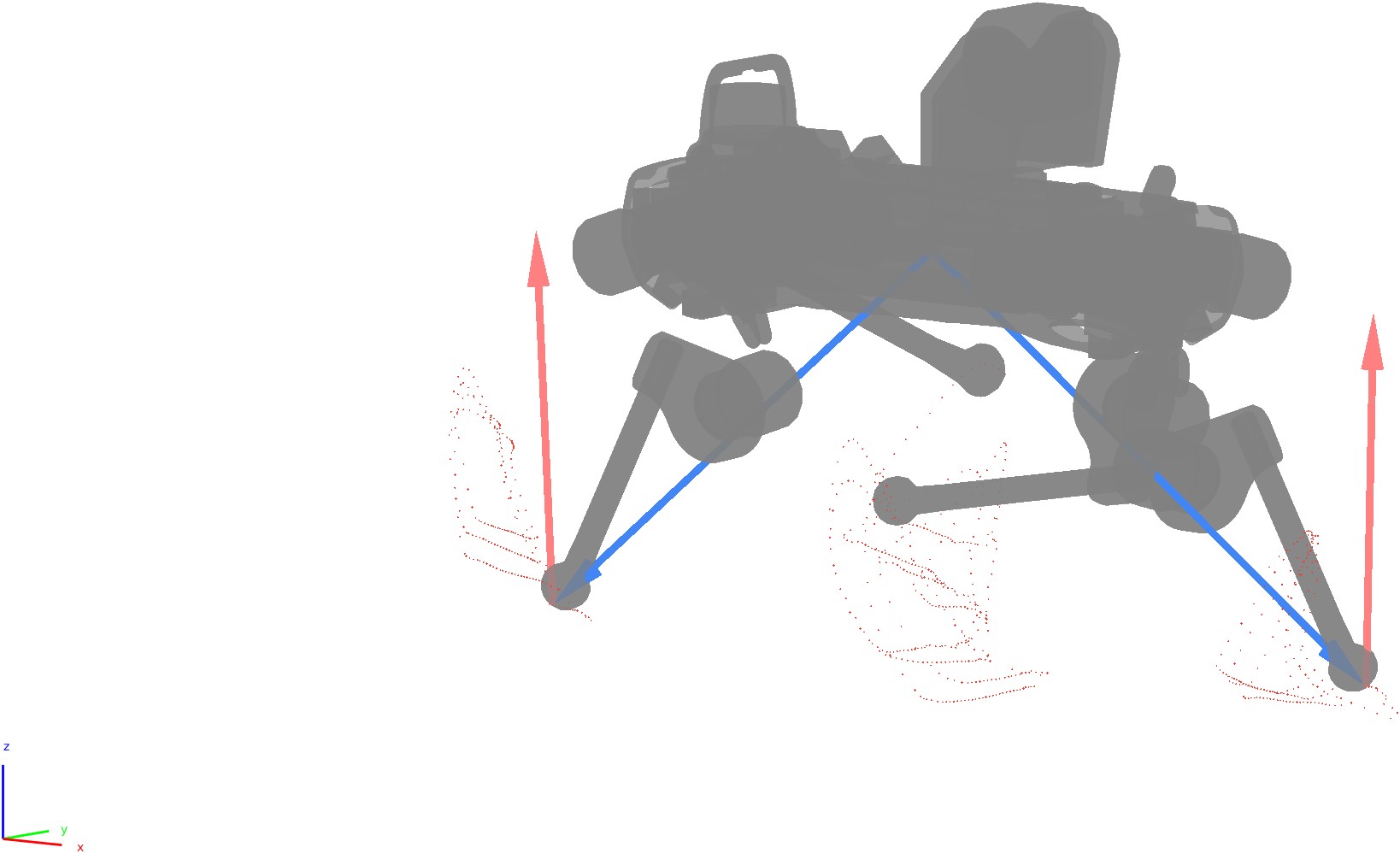}
}
\subfloat[\label{fig:subdivision}]{
    \includegraphics[
        scale=0.1,
        trim=480 160 0 0, 
        clip
    ]{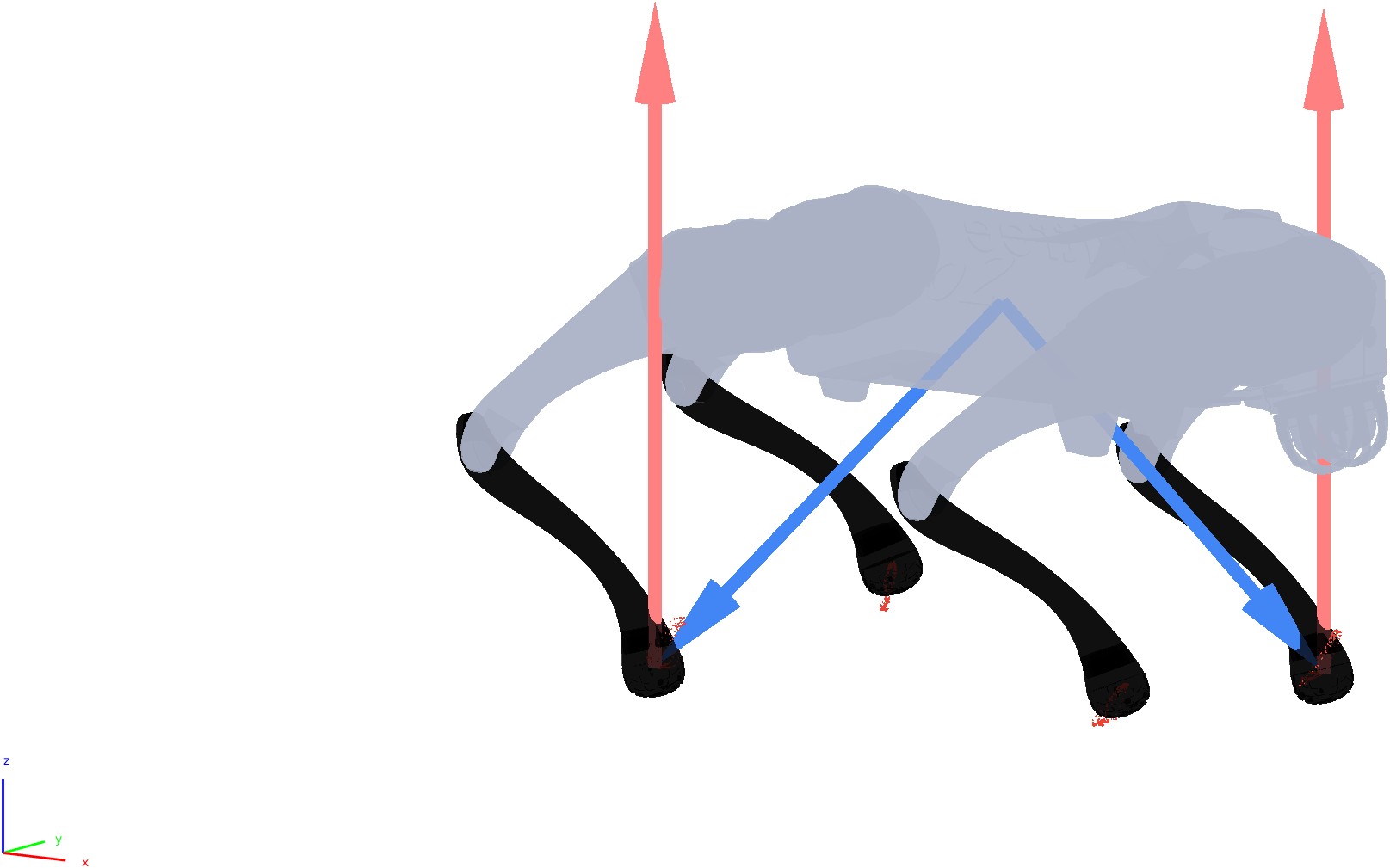}
}
 \caption{ Trajectories, GRFs and relative position of the feet in contact with two robots: a) AnymalD and b) Unitree Go2. } 
 \label{fig:robot_debug}
\label{dataaccel}
\end{figure}
This class is a wrapper for the MATLAB function \path{importrobot}, specialized in
features related to quadruped robots. It depends on the URDF of the robot, which defines the kinematic tree of the robot. Through this class we can obtain the forward kinematics of each foot $f$, $\mathcal{K}_f:\mathbb{S}^{3}\to \mathbb{R}^{3}$ and its Jacobian $J_{\mathcal{K},f}:\mathbb{R}^{3} \to  \mathbb{R}^{3}$. 
This class also depends on \texttt{ContactManagerBase}, responsible for managing the foot's contact. In this work, we implemented a derived class with the simplest implementation: we assume contact if the $z$ component of the GRF, $\mathbf{f}_f$, is above a threshold $\lambda$ that is a private member in the class. As a result, we have a set $S$ of four binary elements stating which feet are in contact or not. By means of this abstract class, we open possibilities for testing new methods in the future without changing the exposed interface. The main class methods are shown in Table \ref{list_robot_functions}. This class 
is exclusive to MATLAB. In the Python version, these operations are obtained from the QuadrupedPyMPC library.
\begin{table}[h!]
\centering
\rowcolors{2}{gray!15}{white}
\scalebox{0.9} {
\begin{tabular}{|c|c|}
\hline
\textbf{Operation} & \textbf{MATLAB's Code}\\
\hline
$\mathbb{S}^{12} \times \mathbb{Z}^{4} \to \mathbb{R}^{3},\;
(\mathbf{q}, i) \mapsto \mathbf{p}_f^{i}$
&
\texttt{robot.GetFootPosition(q, i)}
\\

$\mathbb{S}^{12} \times \mathbb{Z}^{4} \to \mathbb{R}^{3\times 3},\;
(\mathbf{q}, i) \mapsto \mathbf{J}_f^{i}$
&
\texttt{robot.GetFootJacobian(q, i)}
\\

$\mathbb{R}^{4} \to \mathbb{Z}_2^{4},\;
\{f_z^{1}, f_z^{2}, f_z^{3}, f_z^{4}\} \mapsto S$
&
\texttt{robot.GetFeetContact(f)}
\\

\begin{tabular}[c]{@{}l@{}}
$\mathbb{R}^{4\times 12} \to \mathbb{R}^{3\times 4},$ 
$(\boldsymbol{\tau}, \mathbf{q}, \dot{\mathbf{q}}, \ddot{\mathbf{q}})$ \\
$\mapsto \{f_z^{1}, f_z^{2}, f_z^{3}, f_z^{4}\}$
\end{tabular}
&
\texttt{robot.GetGRF(t, q, qd, qdd)}
\\
\hline
\end{tabular}
}
\caption{Main operations associated to a robot in Chalito.}
\label{list_robot_functions}
\vspace{-2mm}
\end{table}

This class is also derived from the \path{Scene} class, which is the manager of the canvas. We can render three representations of the robot: 2D, box 3D, and realistic. 
The realistic representation is especially useful for debugging leg ordering, forward kinematics, estimated GRFs and contact, as shown in Fig. \ref{fig:robot_debug}. 
The other representations are useful for debugging the filter's output.
\subsection{Filter interface}
\label{sec:filter_interface}
In this work, we consider a robot equipped with an IMU mounted on its base, joint encoders in its legs, and force sensors in its feet. 
Based on these sensing modalities, different measurements can be derived to estimate the robot’s state. A summary of the considered measurements is provided in Table~\ref{tab:base_estimation}.
\begin{table}[h!]
\centering
\rowcolors{2}{gray!15}{white}
\begin{tabular}{|>{\raggedright\arraybackslash}p{3cm}|p{4.8cm}|}
\hline
\textbf{Observable} & \textbf{Observation model} \\ \hline
Base displacement (\texttt{InertialOdometry}) & Obtained by integrating IMU measurements \\
Base displacement (\texttt{LegKinematics}) & Computed from the change in the relative position of the base centroid to a static, known foot position as defined in \cite{bloesch2012,Hartley2019ContactaidedIE}  \\
Base velocity (\texttt{BaseVelocity}) & Equal to the average of foot velocities during contact, as defined in \cite{bloesch2013} \\
\hline
\end{tabular}
\caption{Observation models considered in this work.}
\label{tab:base_estimation}
\vspace{-2mm}
\end{table}

We define as state variables of the robot at instant $t_i$ the orientation $\mathbf{R}_i \in  \mathrm{SO}(3)$, 
base position $\mathbf{p}_i \in \mathbb{R}^{3}$, velocity $\mathbf{v}_i \in \mathbb{R}^{3}$, 
position of the feet in contact $\mathbf{p}_f^{k} \in  \mathbb{R}^{3}$, $k = \{1,\ldots,N\}$, $N \leq 4$ and 
IMU biases $\mathbf{b}_{g,i}, \mathbf{b}_{a,i} \in \mathbb{R}^{3}$. All variables except IMU biases are represented in the world frame.

As a motivation for the filter and manifold interfaces proposed in this work, we decided to embed the state variables into an element of two different Lie groups. 
The first one is the group $\mathrm{SO}(3) \times \mathbb{R}^{12 + 3N}$:
\begin{equation}
\label{eq:qekf}
\boldsymbol{x}_i := \{\mathbf{R}_i, \mathbf{v}_i, \mathbf{p}_i, \mathbf{p}_f^{1}, \ldots, \mathbf{p}_f^{N}, \mathbf{b}_{g,i}, \mathbf{b}_{a,i}\},
\end{equation}
while the second is $\mathrm{SE}_{2+N}(3) \times  \mathbb{R}^{6}$:
\begin{equation}
\label{eq:iekf_state}
\mathcal{X}_i := (\mathbf{R}_i, \mathbf{x}_i) \times [\mathbf{b}_{g,i}, \mathbf{b}_{a,i}],\, \mathbf{x}_i = [\mathbf{v}_i, \mathbf{p}_i, \mathbf{p}_f^{1}, \ldots, \mathbf{p}_f^{N}]^{T}.
\end{equation}

While the composition rule for the first group is trivial (it is just the composition of each element), the composition rule for $\mathrm{SE}_{2+N}(3)$ is given by \cite{barrau2022}:
\begin{equation}
(\mathbf{R}_i, \mathbf{x}_i) \circ (\mathbf{R}_j, \mathbf{x}_j) := (\mathbf{R}_i\mathbf{R}_j, \mathbf{x}_i + \mathbf{R}_i * \mathbf{x}_j),
\end{equation}
where $(*)$ denotes a term-by-term multiplication, e.g, $\mathbf{R} * [\mathbf{x}_1, \ldots, \mathbf{x}_n] := [\mathbf{R}\mathbf{x}_1, \ldots, \mathbf{R} \mathbf{x}_n ]$.

We denote the filter based on $\boldsymbol{x}_i$ as the $\mathrm{SO}(3)$-Extended Error State Kalman Filter ($\mathrm{SO}(3)$-EKF).
This filter is equivalent to the one proposed in \cite{bloesch2012}, although they used the 
unit quaternion group to parametrize the $\mathrm{SO}(3)$. The filter based on $\mathcal{X}_i$ is referred to as the Invariant Extended Error State Kalman Filter (IEKF), owing to the invariant properties of the underlying Lie group structure highlighted in \cite{barrau2015, barrau2022}. In both formulations, the error between the nominal and true states is propagated. 
In this work, a right-invariant error formulation is adopted, whereby the estimation error is represented in the Lie algebra and expressed in inertial (world-frame) coordinates. 
Although a left-invariant formulation could be obtained by adapting the right-invariant one, such an extension is not yet supported in our toolbox. 
The initial nominal state in both filters is defined as: 
\begin{equation}
\label{eq:error_equation}
\begin{aligned}
\bar{\mathcal{X}}_0 := (-\boldsymbol{\xi}_0) \oplus \mathcal{X}_0,
\boldsymbol{\xi}_0 \sim \mathcal{N}(\mathbf{0}_{15 + 3N}, \mathbf{P}_i),\\
 \bar{{\boldsymbol{x}}}_0 := (-\delta \boldsymbol{x}_0) \oplus \boldsymbol{x}_0,
\delta \boldsymbol{x}_0
\sim \mathcal{N}(\mathbf{0}_{15 + 3N}, \mathbf{\Sigma}_i).
\end{aligned}
\end{equation}
Notice that $\delta \boldsymbol{x}_0 \approx \mathbf{J}\boldsymbol{\xi}_0$, where $\mathbf{J}$ was given in \cite[Eq. 76]{santana2026}.
This linear relation is useful when performing Monte Carlo simulations, as it allows one to convert a sample from the initial distribution of the IEKF to a sample from the initial distribution of the $\mathrm{SO}(3)$-EKF.

In the following sections we describe how the prediction and update steps can be implemented using the proposed Filter interface.
\subsubsection{Prediction Step}
The prediction step is obtained by integrating the IMU measurements, i.e., angular velocity $\tilde{\boldsymbol{\omega}}_{I,i}$ 
and linear acceleration $\tilde{\mathbf{a}}_{I,i}$. We denote with the subscript $I$ all quantities in the IMU frame. 
We assume the measurements are piecewise constant in the interval 
$[t_i, t_i + \Delta t]$ and that they are corrupted by additive white Gaussian noise and biases modeled as zero-mean random walks:
\begin{equation}
\begin{aligned}
\label{eq:imu_biases}
  \tilde{\boldsymbol{\omega}}_{I,i} = \boldsymbol{\omega}_{I,i} + \mathbf{w}_{g,i} + \mathbf{b}_{g,i}, \mathbf{w}_{g,i} 
  \sim \mathcal{N}(\mathbf{0}_{3,1}, \mathbf{Q}_g), \\
  \tilde{\mathbf{a}}_{I,i} = \mathbf{a}_{I,i} + \mathbf{w}_{a,i} + \mathbf{b}_{a,i}, \mathbf{w}_{a,i} 
  \sim \mathcal{N}(\mathbf{0}_{3,1}, \mathbf{Q}_a), \\
  \mathbf{b}_{g,i + 1} = \mathbf{b}_{g,i} + \mathbf{w}_{bg,i}\Delta t, \mathbf{w}_{bg,i} \sim \mathcal{N}(\mathbf{0}_{3,1}, \mathbf{Q}_{bg}), \\
  \mathbf{b}_{a,i + 1} = \mathbf{b}_{a,i} + \mathbf{w}_{ba,i}\Delta t, \mathbf{w}_{ba,i} \sim \mathcal{N}(\mathbf{0}_{3,1}, \mathbf{Q}_{ba}).
\end{aligned}
\end{equation}
To accommodate some dynamic event, we assume the foot contact position during contact is corrupted by white Gaussian noise in the IMU frame \cite{bloesch2012}:
\begin{equation}
  \label{eq:foot_biases}
        \dot{\mathbf{p}}_f = \mathbf{R}_i\mathbf{w}_f, \mathbf{w}_f \sim \mathcal{N}(\mathbf{0}_{3,1}, \mathbf{Q}_f).
\end{equation}
In this filter formulation, the initial nominal state is propagated through the system dynamics, while the error is propagated through the linearized dynamics. 
The system dynamics for both filters are governed by the IMU and given by \cite{santana2024}:
\begin{equation}
\label{eq:prediction_step}
\begin{aligned}
\bar{\boldsymbol{x}}_{i|i-1} := \mathcal{F}_{i-1}(\bar{\boldsymbol{x}}_{i-1}, 
                                      \tilde{\boldsymbol{\omega}}_{I,i-1},
                                      \tilde{\mathbf{a}}_{I,i-1},
                                      , \Delta t) \approx \\
\begin{bmatrix}
\bar{\mathbf{R}}_{i-1}\mathrm{Exp}(({\tilde{\boldsymbol{\omega}}}_{I,i-1} -\bar{\mathbf{b}}_{g,i-1})\Delta t) \\
\bar{\mathbf{v}}_{i-1} + (\bar{\mathbf{R}}_{i-1} (\tilde{\mathbf{a}}_{I,i-1} - \bar{\mathbf{b}}_{a,i-1}) + \mathbf{g}) \Delta t \\
\bar{\mathbf{p}}_{i-1} + \bar{\mathbf{v}}_{i-1}\Delta t +  (\bar{\mathbf{R}}_{i-1}(\tilde{\mathbf{a}}_{I,i-1} - \bar{\mathbf{b}}_{a,i-1} )+ \mathbf{g})\frac{\Delta t^{2}}{2} \\
\bar{\mathbf{p}}_{f,i-1}\\
\bar{\mathbf{b}}_{g,i-1}\\
\bar{\mathbf{b}}_{a,i-1}
\end{bmatrix},
\end{aligned}
\end{equation}
where we only considered one foot in contact, $\mathbf{g} := [0,0,-9.81]^{T}$, and $\mathrm{Exp}(\boldsymbol{\omega}) := \exp(\boldsymbol{\omega}^{\wedge})$. Equation \eqref{eq:prediction_step} is also used to propagate $\bar{\mathcal{X}}_{i}$.

In the presence of process noise, the approximate covariance propagation for both filters is given by  \cite{bloesch2012,santana2024}:
\begin{equation}
\begin{aligned}
\label{eq:covariance_propagation}
  \mathbf{P}_{i|i-1} \approx \mathbf{A}_{i-1} \mathbf{P}_{i-1} \mathbf{A}_{i-1}^{T} + \mathbf{B}_{i-1} \mathbf{Q}_{i-1} \mathbf{B}_{i-1}^{T}, \\
  \mathbf{\Sigma}_{i|i-1} \approx \bar{\mathbf{A}}_{i-1} \mathbf{\Sigma}_{i-1} \bar{\mathbf{A}}_{i-1}^{T} + \bar{\mathbf{B}}_{i-1} \mathbf{Q}_{i-1} \bar{\mathbf{B}}_{i-1}^{T},
\end{aligned}
\end{equation}
where $\mathbf{Q}_i := \mathrm{diag}(\mathbf{Q}_g, \mathbf{Q}_a, \mathbf{Q}_f, \mathbf{Q}_{bg}, \mathbf{Q}_{ba})$ and the Jacobians 
(\cite[Table 2]{Hartley2019ContactaidedIE} and \cite[Eq. 44]{bloesch2013}) are obtained by linearizing the error 
equation defined in \eqref{eq:error_equation}.
The main methods required in the prediction step were defined in the proposed interface class
and are shown in Table \ref{tab:list_prediction_functions}. The first method is the orchestrator, while the next three implement \eqref{eq:prediction_step} and \eqref{eq:covariance_propagation}, and are provided in the base class. The last two methods compute the Jacobians required by \eqref{eq:covariance_propagation} and are filter-dependent.
The correction in the nominal state is obtained through the innovation and described in the update step.
\begin{table}[h!]
\centering
\rowcolors{2}{gray!15}{white}
\scalebox{0.80} {
\begin{tabular}{|c|c|}
\hline
\textbf{Operation} & \textbf{Interface's Code}\\
\hline
$
\begin{aligned}
&G \times \mathbb{R}^{m} \times \\ 
&\mathbb{R}^{3\times 3} \times \mathbb{R}^{3\times 3} \ldots \times \mathbb{R}^{n \times n},\; \\
&(\bar{\mathcal{X}}_i, \tilde{\mathbf{u}}_{I,i+1}, \mathbf{Q}_a,\ldots, \mathbf{P}_i) \mapsto \\
&(\bar{\mathcal{X}}_{i+1},\mathbf{P}_{i+1})
\end{aligned}$ & {\texttt{filter.PredictionStep(u,[Qa,...],dt)}} \\
\hline
$
\begin{aligned}
&G \times \mathbb{R}^{m} \to  G,\; \\ 
&(\bar{\mathcal{X}}_{i}, \tilde{\mathbf{u}}_{I,i+1}, \Delta t) \mapsto \bar{\mathcal{X}}_{i+1}
\end{aligned}
$ & {\texttt{filter.PropagateState(u,dt)}} \\
\hline
$
\begin{aligned}
&\mathbb{R}^{3\times 3} \times \mathbb{R}^{3\times 3} \ldots  
\to \\
&\mathbf{Q}^{(2+s)(3 \times 3)} ,\;\\
&(\mathbf{Q}_a, \mathbf{Q}_g, \mathbf{Q}_f^{1}, \ldots) \mapsto \mathbf{Q}_i
\end{aligned}$ & {\texttt{filter.GetQ([Qa,...])}} \\
\hline
$
\begin{aligned}
&\mathbb{R}^{n\times n} \times \mathbf{Q}^{(2+s)(3 \times 3)} \\ 
&\times \mathbb{R}^{n \times (2+s)(3)} \times \mathbb{R}^{n \times  n} ,\;\\
&(\mathbf{A}_i, \mathbf{Q}_i, \mathbf{B}_i, \mathbf{P}_i) \mapsto \mathbf{P}_{i+1}
\end{aligned}$ & {\texttt{filter.PropagateCovariance(A,Q,B)}} \\
\hline
$
\begin{aligned}
&G \times \mathbb{R}^{m} \to \mathbb{R}^{n\times n} ,\; \\ 
&(\bar{\mathcal{X}}_i,\tilde{\mathbf{u}}_{I,i+1},\Delta t) \mapsto \mathbf{A}_i
\end{aligned}
$ & \textbf{\texttt{filter.GetA(u,dt)}} \\
\hline
$
\begin{aligned}
&G \times \mathbb{R}^{m} \to \mathbb{R}^{n\times (2+s)(3)} ,\; \\ 
&(\bar{\mathcal{X}}_i,\tilde{\mathbf{u}}_{I,i+1},\Delta t) \mapsto \mathbf{B}_i
\end{aligned}
$ & \textbf{\texttt{filter.GetB(u,dt)}} \\
\hline
\end{tabular}
}
\caption{Prediction-step interface methods. Pure abstract methods in bold; others implemented in the base class. Private members are not method arguments.}
\label{tab:list_prediction_functions}
\end{table}
\subsubsection{Update Step}
Generally, observations are modeled as noisy measurements of the state through a differentiable map
$\mathcal{G}: G \rightarrow \mathbb{R}^m$ \cite{he2021}:
\begin{equation}
\label{eq:observation}
\tilde{\mathbf{y}}_{I,i}
=
\mathcal{G}(\boldsymbol{x}_i)
+
\mathbf{w}_k,
\qquad
\mathbf{w}_k
\sim
\mathcal{N}(\mathbf{0}_{m,1},\mathbf{Q}_k).
\end{equation}
The innovation $\bar{\mathbf{z}}_i$, defined as the difference between the observed and predicted measurements, is linearized with respect to $\delta \boldsymbol{x}$ around the nominal state $\bar{\boldsymbol{x}}_i$, yielding
\begin{equation}
\label{eq:linear_so3}
\bar{\mathbf{z}}_i
:=
\tilde{\mathbf{y}}_{I,i}
-
\mathcal{G}(\bar{\boldsymbol{x}}_i)
=
\bar{\mathbf{H}}(\bar{\boldsymbol{x}}_i)\,
\delta\boldsymbol{x}_i
+
\mathbf{w}_k
+
O(\|\delta\boldsymbol{x}_i\|^2),
\end{equation}
where $\bar{\mathbf{H}}$ denotes the Jacobian of $\mathcal{G}$ with respect to $\delta\boldsymbol{x}$ evaluated at $\bar{\boldsymbol{x}}_i$ \cite{sola2018}. Since the linearization is performed at the nominal rather than the true state, the observability properties depend on $\bar{\mathbf{A}}(\bar{\boldsymbol{x}}_i)$ and $\bar{\mathbf{H}}(\bar{\boldsymbol{x}}_i)$. Consequently, the filter becomes overconfident in unobservable directions and may eventually diverge \cite{huang2008, barrau2015}.

In the IEKF, the observation model is embedded as a group action of the true state~\cite{barrau2022},
\begin{equation}
\label{eq:}
\begin{aligned}
\hat{{\tilde{\mathbf{y}}}}_i := 
\begin{bmatrix}
\tilde{\mathbf{y}}_{i} \\
\mathbf{0}_{d-m,1}  
\end{bmatrix} + \mathbf{d} = \mathcal{X}^{-1}_i \mathbf{d} + \begin{bmatrix}
\mathbf{w}_{f}  \\
\mathbf{0}_{d-m,1} 
\end{bmatrix} =\mathcal{X}^{-1}_i\mathbf{d} + \hat{\mathbf{w}}_{k},
\end{aligned}
\end{equation}
where $\mathbf d \in \mathbb R^d$ is constant. The corresponding innovation is
\begin{equation}
\mathbf z_i
=
\bar{\mathcal X}_i \hat{\tilde{\mathbf y}}_i - \mathbf d
=
\exp(\boldsymbol{\xi}_i^{\wedge})^{-1}\mathbf d
-\mathbf d
+
\bar{\mathcal X}_i \hat{\mathbf w}_f .
\end{equation}
Using the first-order approximation
$
\exp(\boldsymbol{\xi}_i^{\wedge})^{-1}
\approx
\mathbf I-\boldsymbol{\xi}_i^{\wedge},
$
\begin{equation}
\label{eq:linear_iekf}
\mathbf z_i
=
-\boldsymbol{\xi}_i^{\wedge}\mathbf d
+
\bar{\mathcal X}_i \hat{\mathbf w}_f
+
O(\|\boldsymbol{\xi}_i\|^2)
=
\mathbf H_i\boldsymbol{\xi}_i
+
\bar{\mathcal X}_i \hat{\mathbf w}_f
+
O(\|\boldsymbol{\xi}_i\|^2),
\end{equation}
where $\mathbf H_i$ depends only on $\mathbf d$. Therefore, the linearized measurement model is independent of the nominal state, preserving the rank deficiency of the observability matrix through updates and improving filter consistency \cite{barrau2015}.
In the following, we provide $\tilde{\mathbf{y}}_i$, $\bar{\mathbf{H}}_i(\bar{\boldsymbol{x}}_i), \mathbf{H}_i$ and $\mathbf{d}$ for the two measurement functions considered in this work and described in Table \ref{tab:base_estimation}.

\textit{Base displacement with forward kinematics}:
%
\begin{equation}
\label{eq:kinematics}
\begin{aligned}
  \tilde{\mathbf{y}}_{I,i}^{1} := \tilde{\mathcal{K}}(\mathbf{q}) = \mathbf{R}_i^{T}(\mathbf{p}_f^{1} - \mathbf{p}) + \mathbf{w}_{k,i},
\\
\bar{\mathbf{H}}_{i}(\boldsymbol{x}_{i})= 
\begin{bmatrix}
\bar{\mathbf{R}}^{T}(\bar{\mathbf{p}}_{f}^{1} - \bar{\mathbf{p}})^{\wedge} & \mathbf{0}_{3} & -\bar{\mathbf{R}}^{T} & \bar{\mathbf{R}}^{T} & \mathbf{0}_{3,6}
\end{bmatrix}, \\
\mathbf{d}  = 
\begin{bmatrix}
\mathbf{0}_{3,1} \\
0 \\
1 \\
-1 \\
\mathbf{0}_{6,1}
\end{bmatrix},
\mathbf{H}_{i} =
\begin{bmatrix}
\mathbf{0}_{3}  & \mathbf{0}_{3}  & \mathbf{I}_{3}  & -\mathbf{I}_{3}  & \mathbf{0}_{3,6} \\
\mathbf{0}_{9,3}  & \mathbf{0}_{9,3} & \mathbf{0}_{9,3} & \mathbf{0}_{9,3} & \mathbf{0}_{9,3}
\end{bmatrix},
\end{aligned}
\end{equation}
where $\mathbf{q} \in  \mathbb{S}_{12}$ are the joint positions. A scheme similar to the one proposed in \cite{Hartley2019ContactaidedIE}
was implemented to manage new contact positions and to remove old ones.
The first contact position and its covariance propagation follow from $\bar{\mathbf{p}}_i + \bar{\mathbf{R}}_i \tilde{\mathbf{y}}_i$.

\textit{Base velocity with contact feet velocity averaging}:
\begin{equation}
\label{eq:average_velocity}
\begin{aligned}
\tilde{\mathbf{y}}_{I,i}^{2} := \frac{1}{|S|} \sum_{s \in S} 
- \tilde{\boldsymbol{\omega}}^{\wedge}_{i,I} \mathbf{p}_{f,B}^{s} - J_{\mathcal{K},s} \dot{\mathbf{q}}_{s} = \mathbf{R}_{i}^{T} \mathbf{v} + \mathbf{w}_{v,i},\\ 
\bar{\mathbf{H}}_{i}(\bar{\boldsymbol{x}}_{i}) = 
\begin{bmatrix}
\bar{\mathbf{R}}_{i}^{T} \bar{\mathbf{v}}_{i}^{\wedge} & \bar{\mathbf{R}}_{i}^{T} & \mathbf{0}_{3} & \mathbf{0}_{3} & \mathbf{0}_{3}
\end{bmatrix},  \\
\mathbf{d}= 
\begin{bmatrix}
\mathbf{0}_{3,1} \\
-1 \\
0 \\
0 \\
\mathbf{0}_{6,1} \\
\end{bmatrix}, 
\mathbf{H}_{i} =
\begin{bmatrix}
\mathbf{0}_{3}  & \mathbf{I}_{3}  & \mathbf{0}_{3}  & \mathbf{0}_{3}  & \mathbf{0}_{3, 6} \\
\mathbf{0}_{9,3}  & \mathbf{0}_{9,3} & \mathbf{0}_{9,3} & \mathbf{0}_{9,3} & \mathbf{0}_{9,3}
\end{bmatrix},
\end{aligned}
\end{equation}
where $\dot{\mathbf{q}}_{s}$ are the joint velocities of leg $s$. 

Once we find a linear stochastic restriction for the error, 
\eqref{eq:linear_so3} and \eqref{eq:linear_iekf}, the Kalman gain is just a way to enforce this restriction in the observable part of the state. 
For simplicity, we only give the equations for the $\mathrm{SO}(3)$-EKF; the IEKF is analogous \cite{santana2024}:
\begin{equation}
\label{eq:update_cov}
\begin{aligned}
\mathbf{S}_{i} = \bar{\mathbf{H}}_{i} \mathbf{\Sigma}_{i|i-1} \bar{\mathbf{H}}_i^{T} 
+ \mathbf{Q}_{k,i}, \\
\mathbf{K}_{i} = \mathbf{\Sigma}_{i|i-1} \bar{\mathbf{H}}_{i}^{T}\mathbf{S}_{i}^{-1},
\\
\delta \hat{\boldsymbol{x}}_{i} = \mathbf{K}_{i} \bar{\mathbf{z}}_{i}, \\
\mathbf{\Sigma}_{i} = (\mathbf{I}_{18} - \mathbf{K}_{i} \bar{\mathbf{H}}_{i})\mathbf{\Sigma}_{i|i-1}.
\end{aligned}
\end{equation}
Finally, the nominal state is updated with the estimated error:
\begin{equation}
\label{eq:update_state}
\bar{\boldsymbol{x}}_{i} :=  \hat{\boldsymbol{x}}_{i} := 
\delta \hat{\boldsymbol{x}}_{i} \oplus \bar{\boldsymbol{x}}_{i|i-1}.
\end{equation}
The \texttt{FilterBase} class implements the core methods for the filter update step (summarized in Table~\ref{list_filter_functions}). The base class explicitly defines two primary functions: \texttt{filter.\allowbreak{}Update(H,N,z)}, which executes the covariance and state update equations \eqref{eq:update_cov} and \eqref{eq:update_state}; and \texttt{filter.\allowbreak{}ProcessMeasurement(meas)}, which orchestrates the update pipeline based on the measurement type. Specifically, the latter sequentially invokes the filter-dependent routines \texttt{filter.\allowbreak{}Get\allowbreak{}Innovation<Name>}, \texttt{filter.\allowbreak{}Get\allowbreak{}H<Name>}, and \texttt{meas.\allowbreak{}Get\allowbreak{}Measurement\allowbreak{}Covariance} before calling the underlying update function. The input to this pipeline is a set of measurements $\{\tilde{\mathbf{y}}_{I,i}\}$ whose dimension depends dynamically on the number of feet in contact with the ground.
\begin{table}[h!]
\centering
\rowcolors{2}{gray!15}{white}
\scalebox{0.85} {
\begin{tabular}{|c|c|}
\hline
\textbf{Operation} & \textbf{Interface's Code}\\
\hline
$\begin{aligned}
&G \times \mathbb{R}^{m} \times \mathbb{R}^{m} \ldots \to G, \\
&(\bar{\mathcal{X}}_i, \{\tilde{\mathbf{y}}_i\}) \mapsto \bar{\mathcal{X}}_{i+1}
\end{aligned}$ 
                   & \texttt{filter.ProcessMeasurement([y1,...])} \\
$\begin{aligned}
&G \times \mathbb{R}^{d \times n} \times \\
&\mathbb{R}^{d \times n} \times  \mathbb{R}^{d} \to G, \\
&(\bar{\mathcal{X}}_{i}, \mathbf{H}_i, \mathbf{N}_i, \mathbf{z}_i) \mapsto \bar{\mathcal{X}}_{i+1}
\end{aligned}$ & \texttt{filter.Update(H,N,z)} \\
$
\begin{aligned}
&G \times \mathbb{R}^{m} \times  \ldots \to  \mathbb{R}^{m^{\ldots}},\;\\ 
&(\bar{\mathcal{X}}_i, \{\tilde{\mathbf{y}}_i\}) \mapsto \mathbf{z}_i
\end{aligned}$
& \textbf{\texttt{filter.GetInnovation<Name>([y1,...])}} 
\\
$G \to \mathbb{R}^{d \times n},\; \bar{\mathcal{X}}_i \mapsto {\mathbf{H}}_i$ & \textbf{\texttt{filter.GetH<Name>()}} \\
\hline
\end{tabular}
}
\caption{Update-step interface methods. \texttt{<Name>} should be replaced with the measurement (e.g., \texttt{LegKinematics}).
}
\label{list_filter_functions}
\vspace{-8mm}
\end{table}
\subsection{Measurement interface}
A measurement provides information that constrains the system state, either at a single time instant or between two consecutive time instants. Such a relationship can be represented either by a motion model, which relates the current state to a previous state, or by a measurement model, which relates state variables at the same time instant. We provide the method \texttt{GetIntegration(Xi)} (e.g., \eqref{eq:prediction_step}) to represent the output of a motion model, whereas \texttt{GetMeasurement()} represents the output of a measurement model. In the proposed library, the IMU (\texttt{InertialOdometry}) is the only measurement modeled through a motion model, while \texttt{LegKinematics} \eqref{eq:kinematics} and \texttt{BaseVelocity} \eqref{eq:average_velocity} are represented through measurement models.
Measurements are subject to two distinct sources of uncertainty: process noise and measurement noise. Process noise models stochastic disturbances affecting the system dynamics, such as IMU and foot-contact biases (see \eqref{eq:imu_biases} and \eqref{eq:foot_biases}). Measurement noise accounts for sensor inaccuracies and observation errors, including $\mathbf{w}_k$ and $\mathbf{w}_v$ in the kinematics and base-velocity models, and $\mathbf{w}_a$ and $\mathbf{w}_g$ in the accelerometer and gyroscope measurements, respectively. To explicitly represent these uncertainty sources, the measurement interface provides the methods \texttt{GetProcessNoise()} and \texttt{GetMeasurementNoise()}.
The \texttt{MeasurementBase} class defines a unified interface for all measurement types. Its constructor receives an instance of the \texttt{QuadrupedRobot} class, the covariance matrices associated with the measurement, a timestamp, and a vector $\mathbf{u}$ containing the raw sensor outputs (e.g., encoder readings). These quantities constitute the complete set of information required to implement the methods defined by the interface presented in Table~\ref{list_measurements_table}.
\begin{table}[h!]
\centering
\rowcolors{2}{gray!15}{white}
\scalebox{0.9} {
\begin{tabular}{|c|c|}
\hline
\textbf{Operation} & \textbf{Interface's Code}\\
\hline
$\mathbb{R}^{k} \to  \mathbb{R}^{m},\; \tilde{\mathbf{u}}_i \mapsto \tilde{\mathbf{y}}_i$ & \texttt{meas.GetMeasurement()} \\
\hline
$
\begin{aligned}
G \times \mathbb{R}^{m} \to  G,\; \\
(\bar{\mathcal{X}}_i, \tilde{\mathbf{u}}_{i+1}) \mapsto \bar{\mathcal{X}}_i
\end{aligned}
$ & \texttt{meas.GetIntegration(Xi)} \\
\hline
$\mathbf{Q}_i$ & \texttt{meas.GetProcessCovariance()} \\
\hline
$\mathbf{Q}_{k,i}$ & \texttt{meas.GetMeasurementCovariance()} \\
\hline
"measurement name" & \texttt{meas.GetMeasurementName()} \\
\hline
\end{tabular}
}
\caption{Main methods in the \texttt{MeasurementBase} interface.}
\label{list_measurements_table}
\vspace{-8.0mm}
\end{table}

\subsection{Config file}
\label{sec:config_file}
The primary input for using Chalito is a set of synchronized measurements, which can be obtained either by simulating prescribed dynamics with access to ground-truth values or by collecting data from a simulator or real-world experiments. The latter two cases are supported through a configuration file in the \texttt{json} format that specifies the filter name, initial state, the initial covariance, the measurement covariances, and three paths: one for the URDF file, one for the sensor measurements, and an optional one for ground-truth data. Sensor data must be provided as a \texttt{csv} file following the structure detailed in Table \ref{tab:proprioceptive_data}, that contains only proprioceptive information. While ground-truth data, if available, should comprise: timestamp, position, orientation in unit quaternion, and base velocity. The library already has a substantial collection of distinct datasets, including all of Table \ref{tab:quadruped_datasets}.
\begin{table}[h]
\centering
\rowcolors{1}{gray!15}{white}
\begin{tabular}{|c|c|c|c|c|c|c|c|c|c|}
\hline
$t_i$ & $\tilde{\mathbf{a}}_i$ & $\tilde{\boldsymbol{\omega}}_i$ & $\mathbf{q}$ & $\dot{\mathbf{q}}$ & $\mathbf{f}_i^{1}$ & $\mathbf{f}_i^{2}$ & $\mathbf{f}_i^{3}$ & $\mathbf{f}_i^{4}$ \\
\hline
1 & 1:3 & 1:3 & 1:12 & 1:12 & 1:3 & 1:3 & 1:3 & 1:3 \\
\hline
\end{tabular}
\caption{Structure for the input data file: timestamp, linear acceleration, 
angular velocity, joint positions, joint velocities, ground reaction forces.}
\label{tab:proprioceptive_data}
\end{table}
\section{RESULTS}
\label{sec:results}
In this section, we demonstrate the library’s functionalities through three experiments. In the first experiment, 
we compare the time of convergence between IEKF and $\mathrm{SO}(3)$-EKF to a prescribed state using the PyChalito library.
In the second experiment, we perform a Monte Carlo simulation using the Chalito library. In this simulation, we compare 
the average error of the observable state: relative velocity in the base frame and gravity direction. Moreover, we also compute the average Normalized Estimation Error Squared (NEES), 
defined as follows:
\begin{equation}
\mathrm{NEES}_{i} = \frac{1}{K} \sum_{k=1}^{K} (\hat{\boldsymbol{\xi}}_{i}^{k})^{T} (\mathbf{P}_{i})^{-1} \hat{\boldsymbol{\xi}}_{i}^{k}, \\
\end{equation}
where we should use $\delta \hat{\boldsymbol{x}}_i^{k}$ for $\mathrm{SO}(3)\times \mathbb{R}^{6}$ and $K$ is the number of realizations.
In the last experiment, we estimate the robot trajectory using both filters again across six different real-world public datasets spanning five different robots.
\vspace{-4mm}

\subsection{State Estimation in a MuJoCo simulation under large initial error}
In this experiment, we evaluate the performance of the two filters with the Unitree AlienGo robot walking on a ramp. The initial state is perturbed with a large error, resulting in the following 
nominal initial state:
\begin{equation}
\bar{\mathbf{R}}_{0} = 
\begin{bmatrix}
0 & 0 & -1 \\
0 & 1 & 0 \\
1 & 0 &  0
\end{bmatrix}, 
\bar{\mathbf{v}}_{0} = \begin{bmatrix}
2.5 \\
2.0 \\
3.0
\end{bmatrix},
\bar{\mathbf{p}}_{0} = 
\begin{bmatrix}
-0.5 \\
0.0 \\
0.2
\end{bmatrix}.
\end{equation}
The IMU measurements are corrupted with covariance $\mathbf{Q}_i = 10^{-2}\mathbf{I}_6$, without biases. For the update step, we only considered the base velocity measurement, with covariance
estimated from a preliminary run and given by
$\mathbf{w}_{v,i} \sim \mathcal{N}(\mathbf{0}_{3,1}, \mathrm{diag}(7.84, 4.00, 25.00) \times 10^{-4})$.

\begin{figure}[t!]
  \centering
  \includegraphics[scale=0.19]{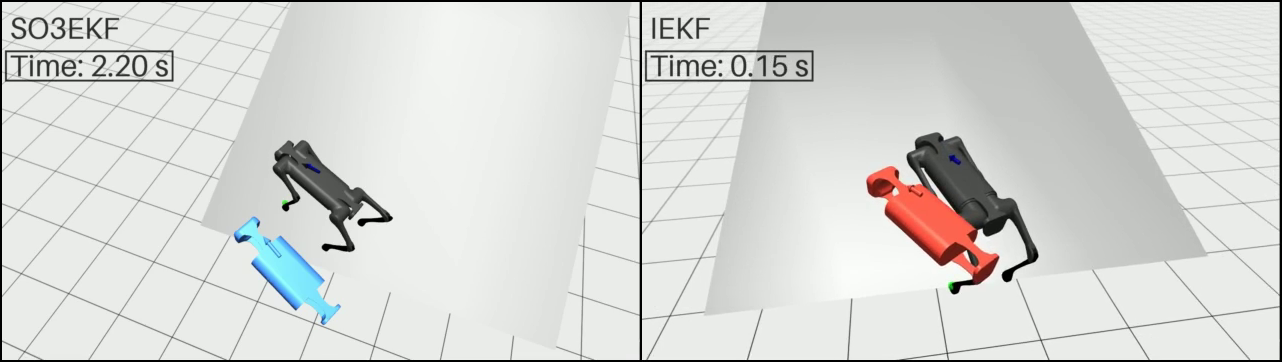}
\caption{
	Time of convergence of the two filters considered in this work IEKF and $\mathrm{SO}(3)$-EKF, with the same initial state, process and measurement white noise. }
\label{fig:mujoco_simulation}
\vspace{-4mm}
\end{figure}

The performance is assessed in terms of convergence time to reach predefined error thresholds of $0.05\, \mathrm{m/s}$ for base velocity, and
$0.015\, \mathrm{rad}$ for the gravity direction. As shown in Fig.~\ref{fig:mujoco_simulation}, the IEKF converged $15$ times faster than $\mathrm{SO}(3)$-EKF.
This demonstrates that the invariant filtering approach significantly improves convergence speed in the presence of large initial errors.
\vspace{-4mm}
\subsection{Monte Carlo simulation with synthetic data}
\begin{figure*}[t]
    \centering
        \includegraphics[scale=0.425]{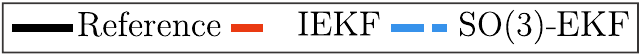}
    \subfloat[\protect\label{fig:exp_1_error_vel}]{
        \includegraphics[width=0.28\textwidth]{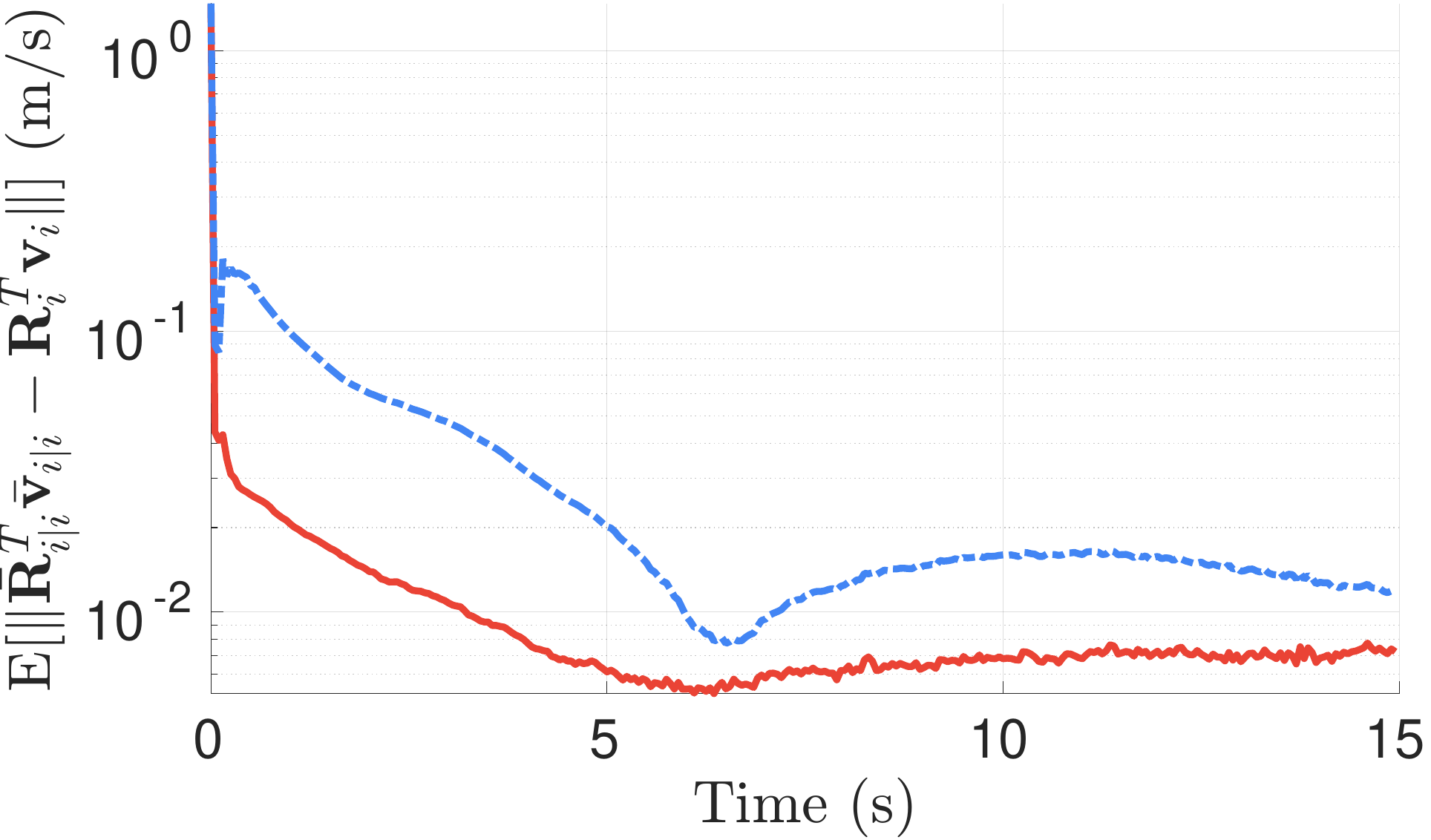}
    }
    \subfloat[\protect\label{fig:exp_1_error_u}]{
        \includegraphics[width=0.28\textwidth]{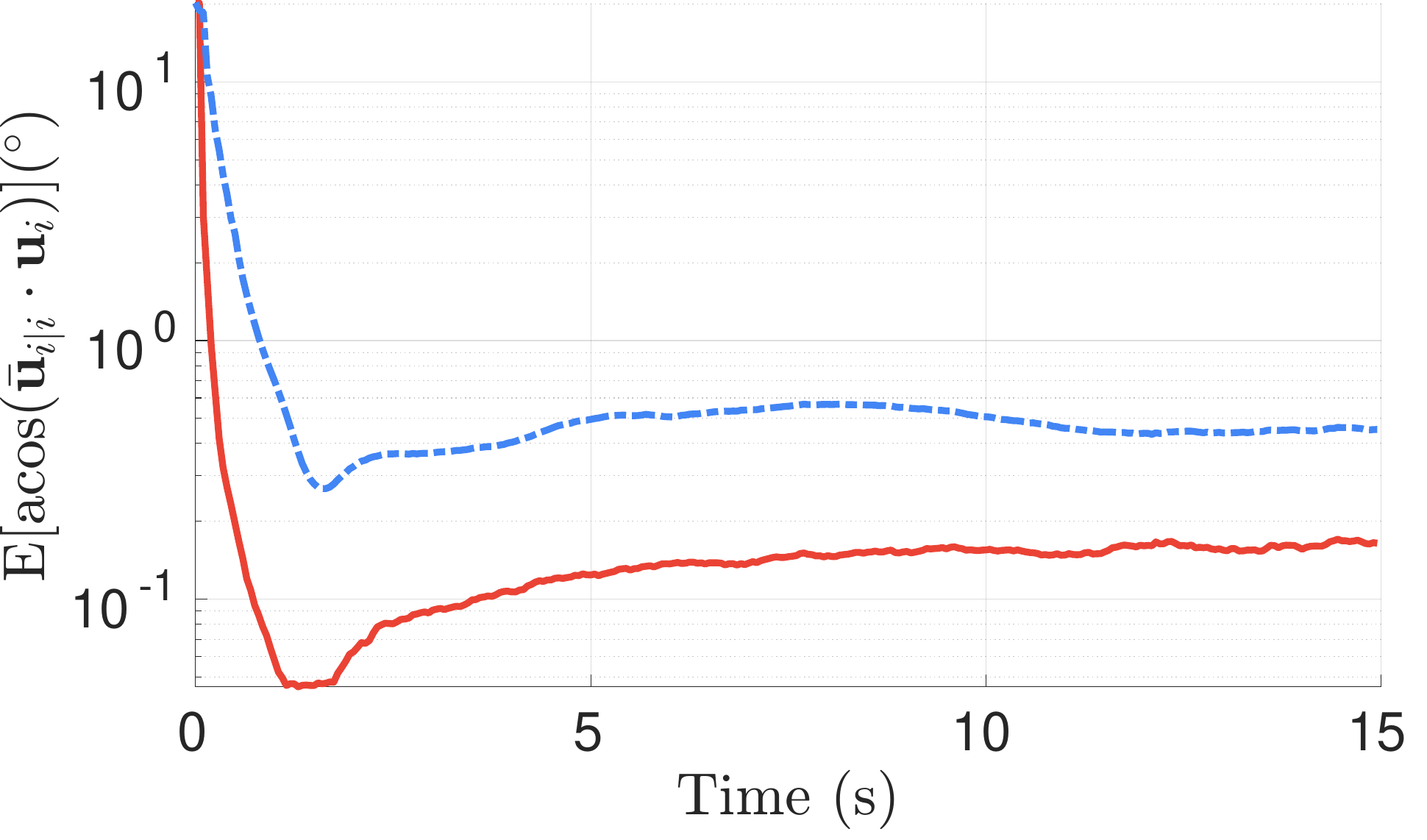}
    }
    \subfloat[\protect\label{fig:exp_1_nees}]{
        \includegraphics[width=0.28\textwidth]{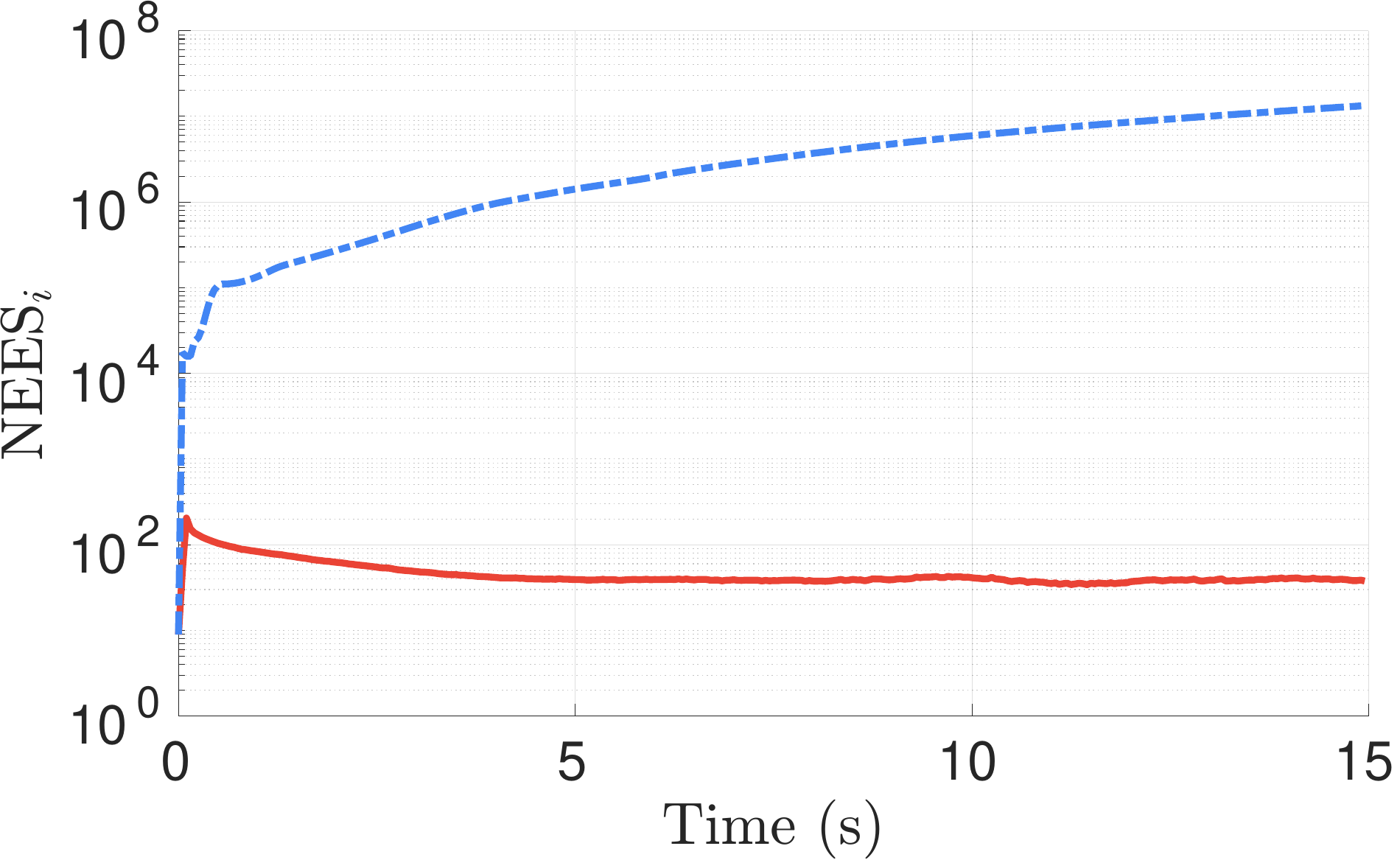}
    }
    \hfill
    \subfloat[\protect\label{fig:exp_position}]{
      \includegraphics[width=0.28\textwidth]{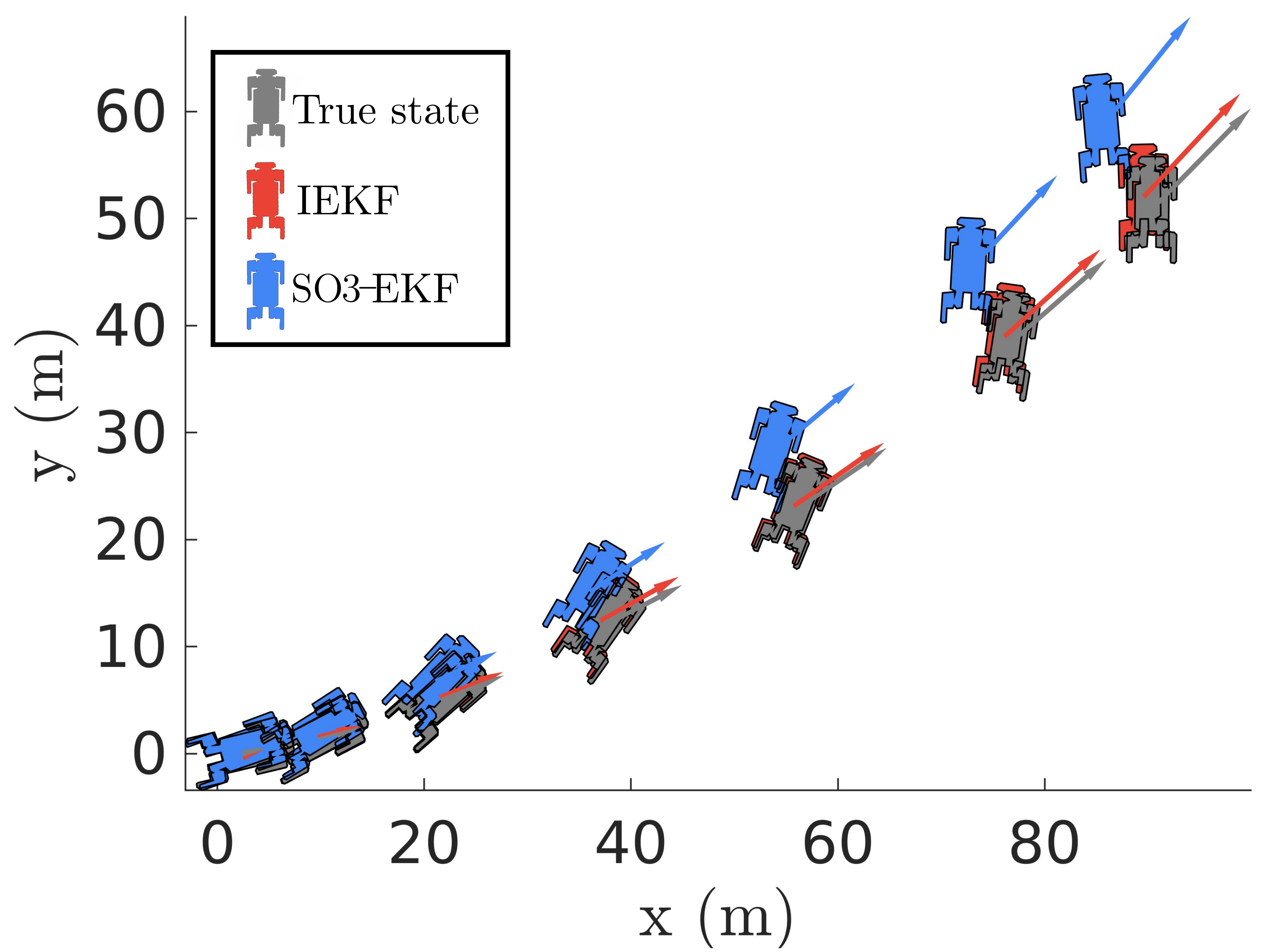}
    }
    \subfloat[\protect\label{fig:exp_pitch}]{
      \includegraphics[width=0.28\textwidth]{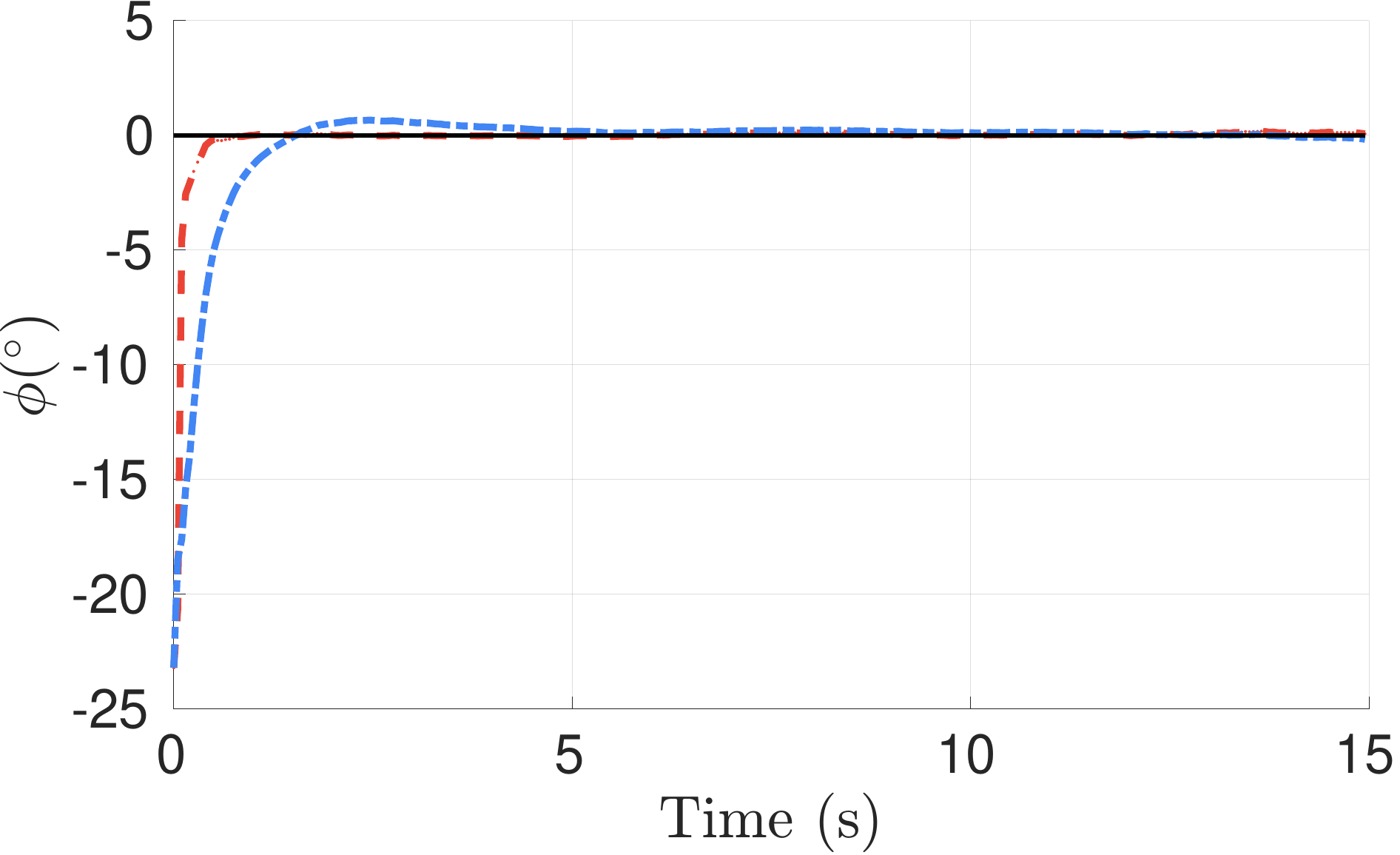}
    }
    \subfloat[\protect\label{fig:exp_gyro_y}]{
      \includegraphics[width=0.28\textwidth]{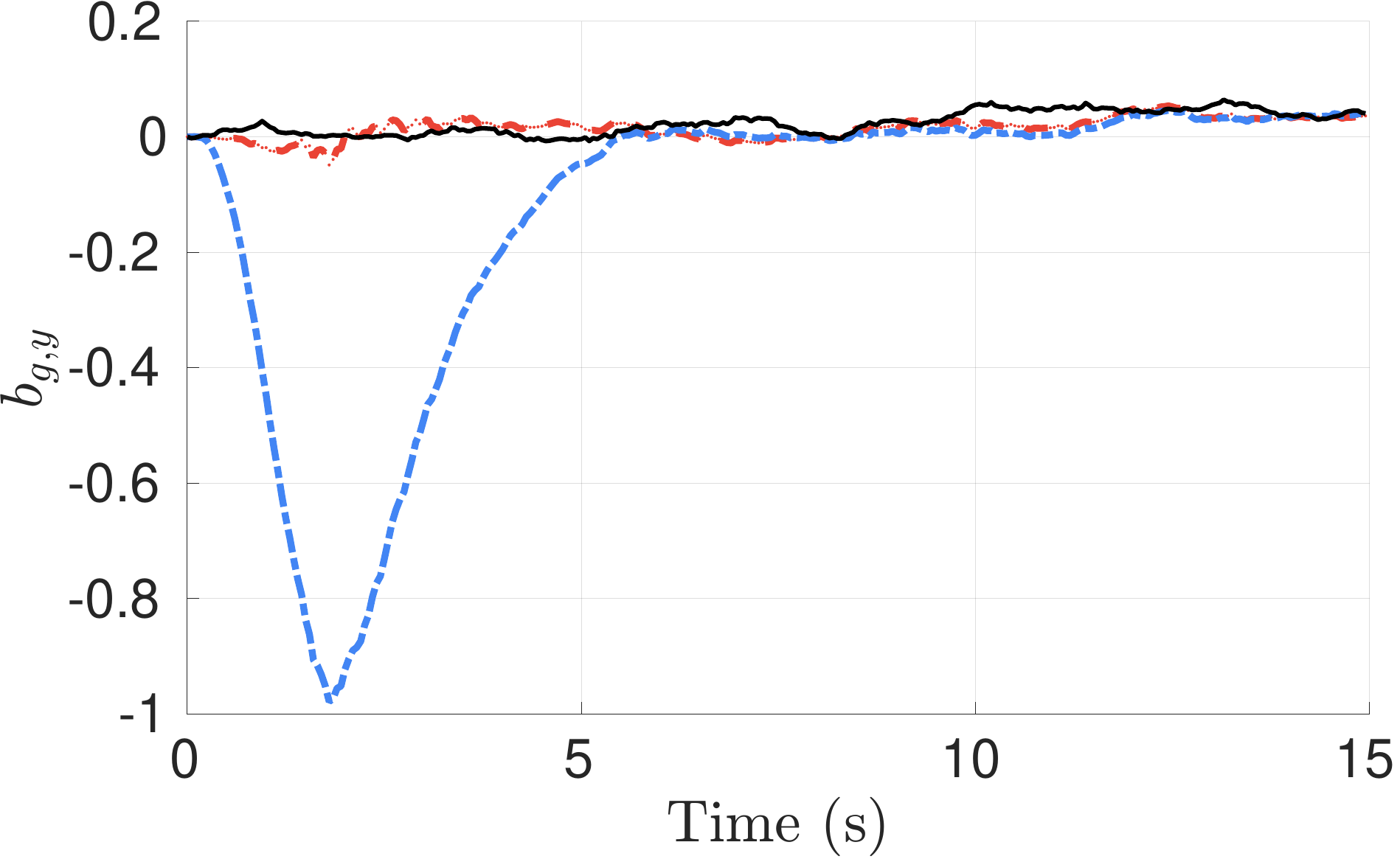}
    }
    \caption{Results of the numerical Monte Carlo experiment with synthetic data. 
        a) Average normalized error for the velocity in the base frame. 
        b) Average normalized error for the gravity direction.
        c) Average NEES. 
        Estimated d) positions, e) pitch angle and f) gyroscope bias 
        in the $y$-axis in one realization of the Monte Carlo simulation.
    }
    \label{fig:experiment_1}
\end{figure*} 
In this next experiment, we assume that the IMU is measuring a constant linear acceleration and angular velocity.  Moreover,
we only consider the base velocity measurement as the update step, due to the simplicity in simulating it.
We apply the prediction and update steps $m$ times in a row. We repeat this procedure for $K$ realizations.
For the motion to remain in the $xy$ plane, we only assume gyroscope white noise in the $z$-direction (yaw angle).
We also assume white noise acceleration in all directions, and biases in the accelerometer and gyroscope measurements. 
All parameters are described as follows:
\begin{equation}
\label{eq:simulation_noise}
\begin{aligned}
\mathbf{P}_{0} =
\mathrm{diag}\left(
\left(\frac{\pi}{12}\right)^2 \mathbf{I}_3,\,
\mathbf{I}_3,\,
10^{-2}\mathbf{I}_3
\right), \\
\mathbf{w}_{\ell,i}
\sim
\mathcal{N}(\mathbf{0}_{3,1}, \mathbf{Q}_{\ell}),
\qquad
\ell \in \{a,g,b_a,b_g,v\}, \\
\mathbf{Q}_{a}
= 2.5\times10^{-5}\mathbf{I}_3, 
\mathbf{Q}_{g}
= \mathrm{diag}(0,0,9\times10^{-4}), \\
\mathbf{Q}_{b_a}
= 2.5\times10^{-11}\mathbf{I}_3, 
\mathbf{Q}_{b_g}
= 10^{-6}\mathbf{I}_3, \\
\mathbf{Q}_{v}
= \mathrm{diag}(10^{-4},10^{-5},10^{-3}), K = 100, \Delta t = 0.05,\\
\mathbf{a}_{I,i}=[1, 0, 9.81]^T, 
\boldsymbol{\omega}_{I,i}
= \Big[0, 0 ,\frac{\pi/2}{m \Delta t}\Big]^T,
m = 300.
\end{aligned}
\end{equation}
We present the main simulation results in Fig.~\ref{fig:experiment_1}. 
Figures~\ref{fig:exp_1_error_vel} and \ref{fig:exp_1_error_u} show the average norm of the velocity error and gravity direction error, respectively. 
It can be observed that the IEKF converged faster than the $\mathrm{SO}(3)$-EKF.
More specifically, the $\mathrm{SO}(3)$-EKF required $2.5$ times more time to achieve the same base velocity error as the IEKF. To measure accuracy, we computed the Mean Absolute Error 
over the first $5 \, \mathrm{s}$ of the average error. The $\mathrm{SO}(3)$-EKF had an error in the velocity by $172 \%$ and in the gravity direction by $160 \%$ with respect to the IEKF.
Figure~\ref{fig:exp_1_nees} reports the average NEES, which, for this problem, should be close to $15$. Here, we observe the false observability issue that is common in the $\mathrm{SO}(3)$-EKF methodology: the filter becomes overconfident because the observability matrix for this problem is incorrectly modeled as fully observable. 
In contrast, the IEKF remains close to 15 from the start, indicating consistent uncertainty estimates. 
Figures ~\ref{fig:exp_position}, \ref{fig:exp_pitch}, \ref{fig:exp_gyro_y} show a realization for the estimated robot position, pitch angle, and $y$ component of gyroscope bias. It can be observed that the position and yaw are unobservable, while the body-frame velocity, pitch angle, and gyroscope bias are accurately estimated.
Finally, besides the synthetic-data experiments presented here, the library supports loading simulated datasets with ground truth and corrupting them with prescribed noise models, enabling the same convergence and consistency analyses in more realistic scenarios.
\vspace{-7mm}

\subsection{Position estimation on real-world datasets}
\begin{figure*}[t!]
      \centering
        \includegraphics[scale=0.450]{./pictures/real_world_results/legend.pdf}
        \subfloat[\protect\label{fig:grantour}]{
          \includegraphics[scale=0.10]{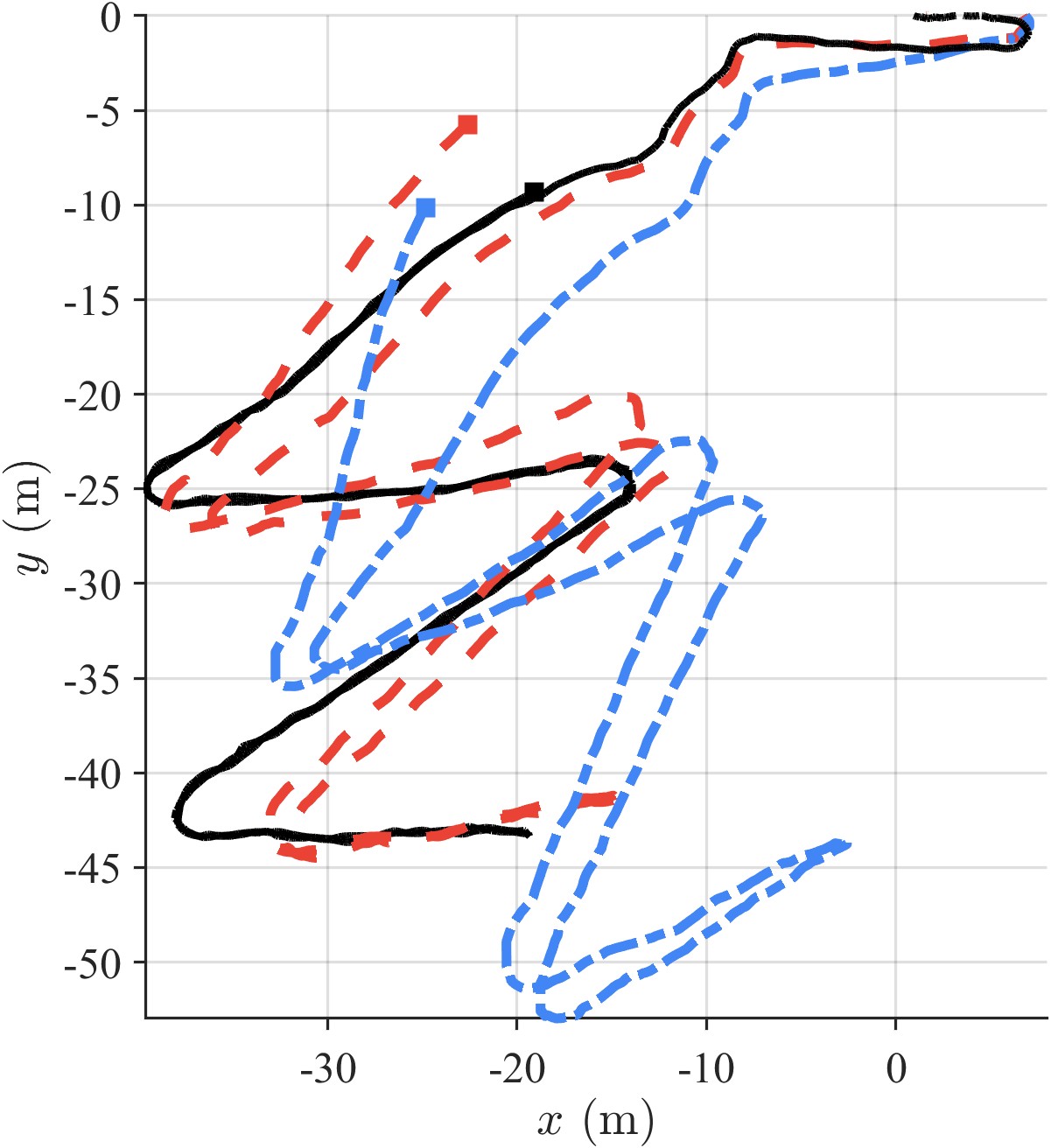}
    }
    \subfloat[\protect\label{fig:cerberus_1}]{
      \includegraphics[scale=0.10]{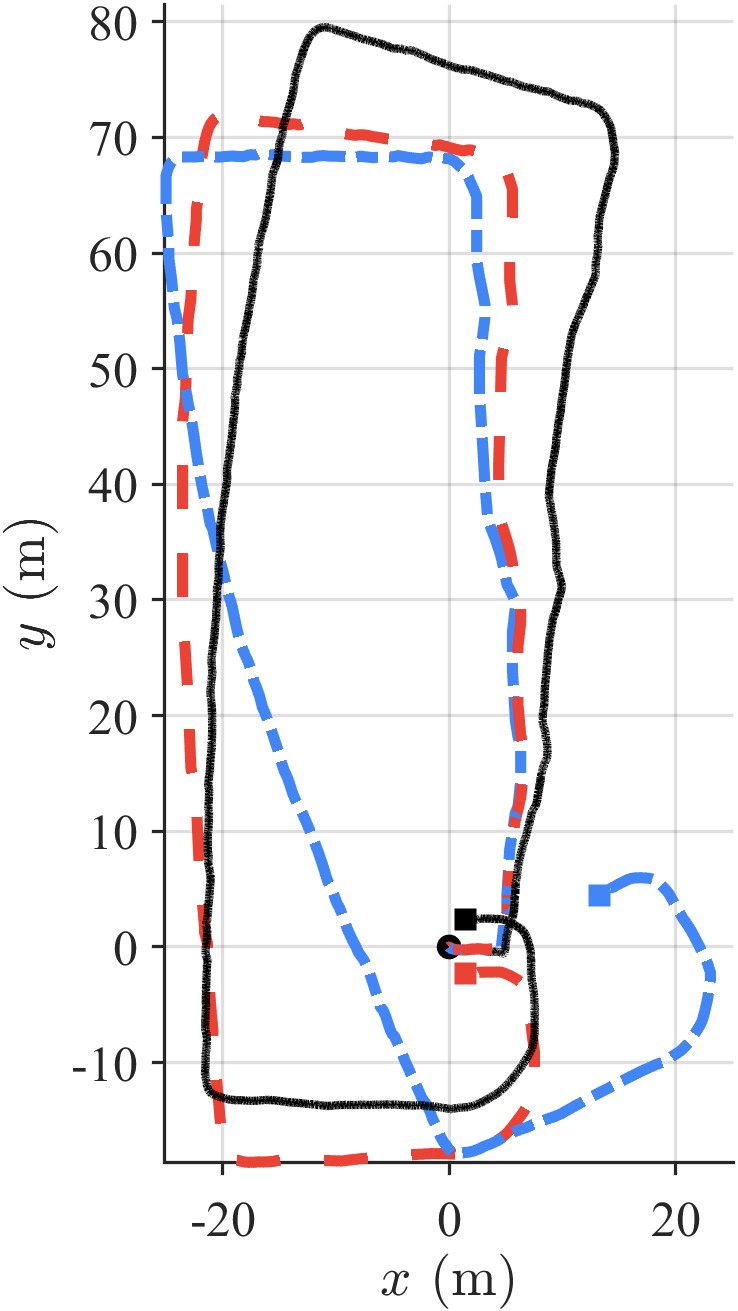}
    }
    \subfloat[\protect\label{fig:pronto}]{
      \includegraphics[scale=0.10]{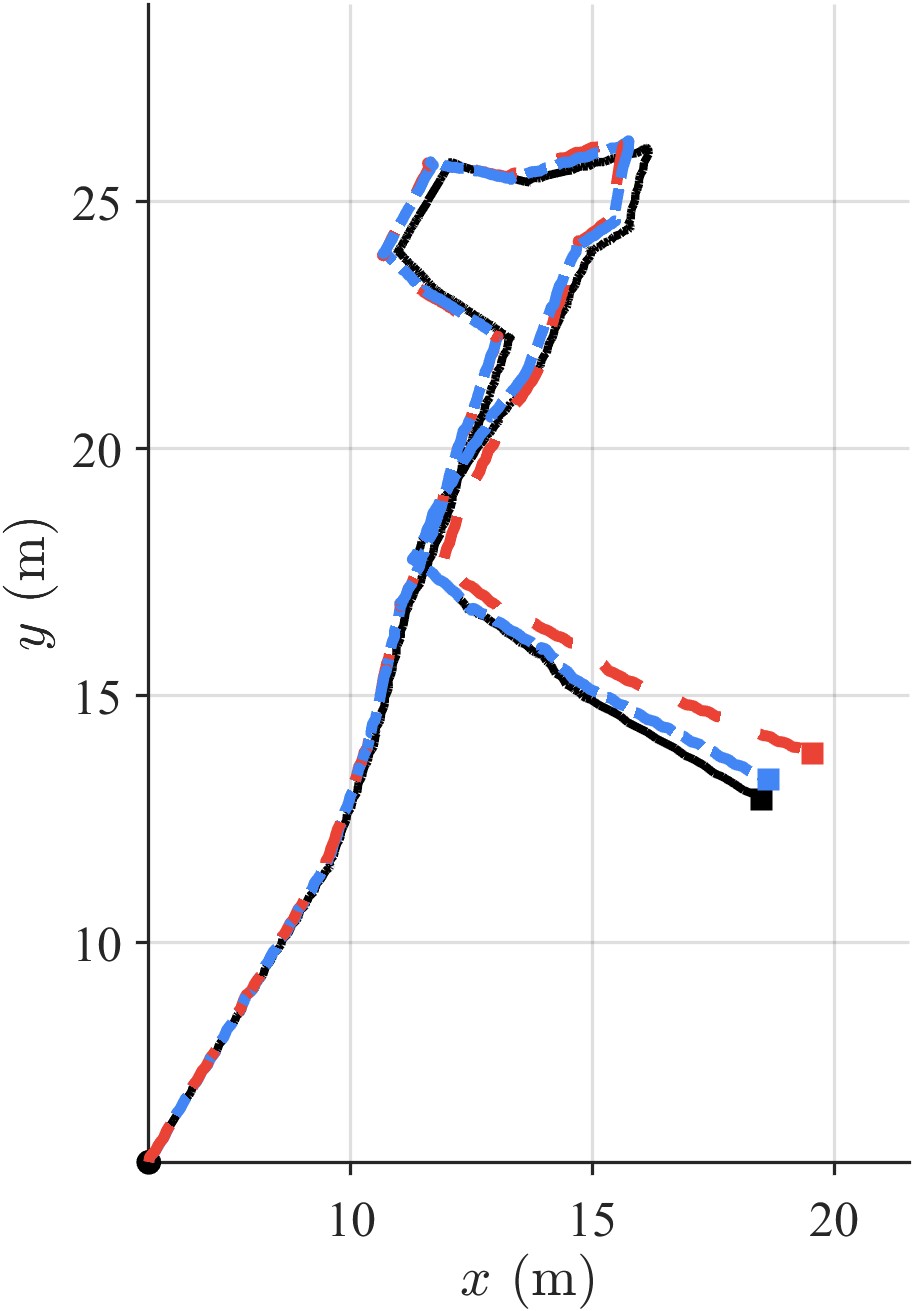}
    }
    \subfloat[\protect\label{fig:legkilo}]{
      \includegraphics[scale=0.10]{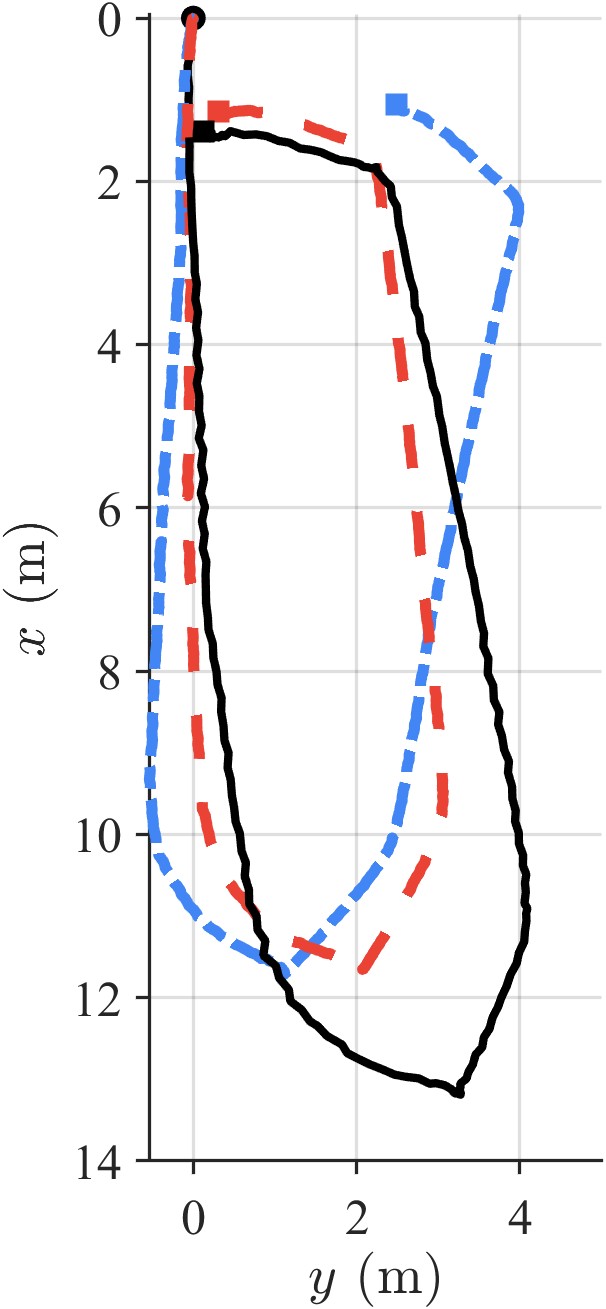} 
    }
   \hfill
   \subfloat[\protect\label{fig:cerberus_2}]{
     \includegraphics[scale=0.10]{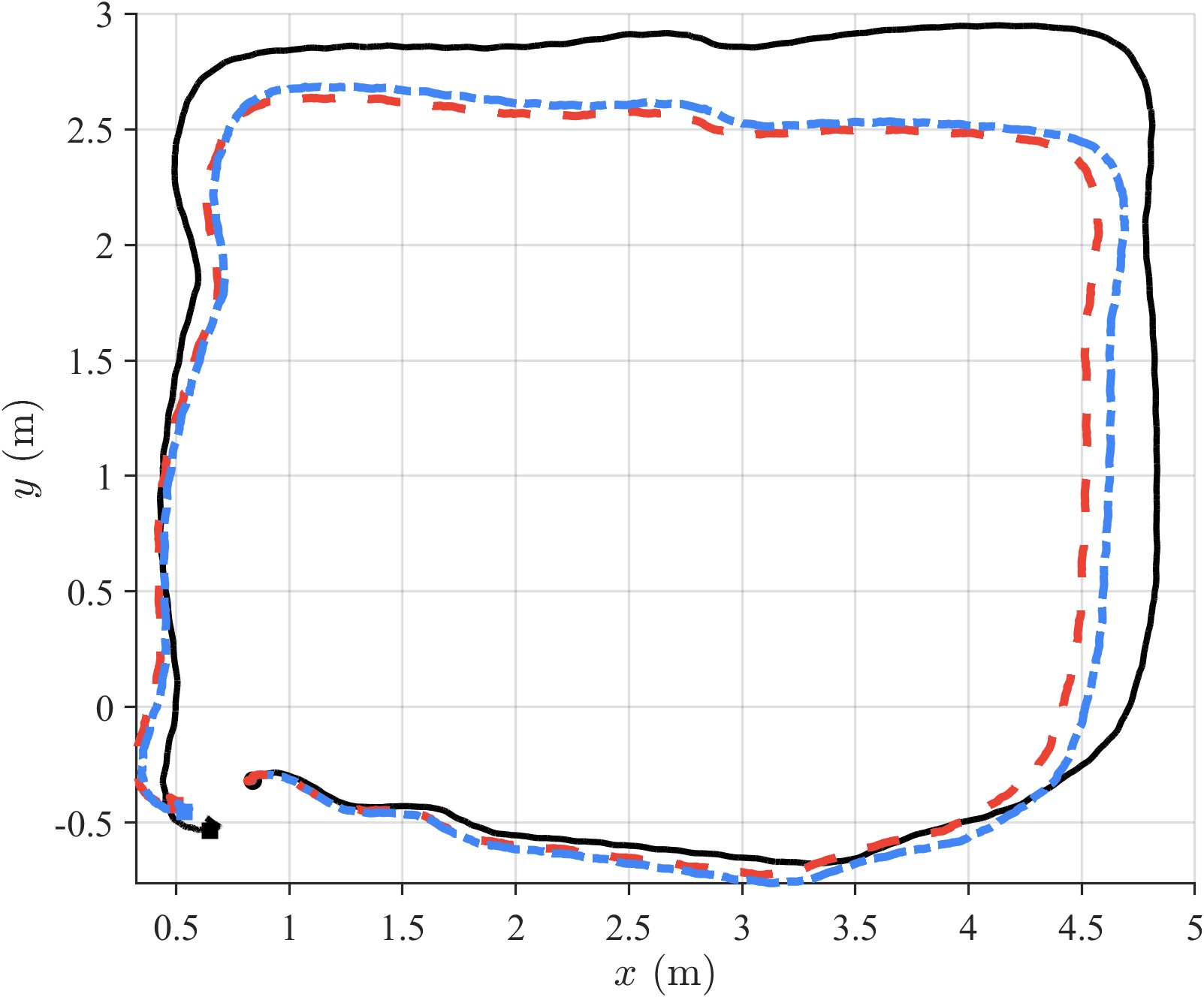}
    }
    \subfloat[\protect\label{fig:dls}]{
      \includegraphics[scale=0.10]{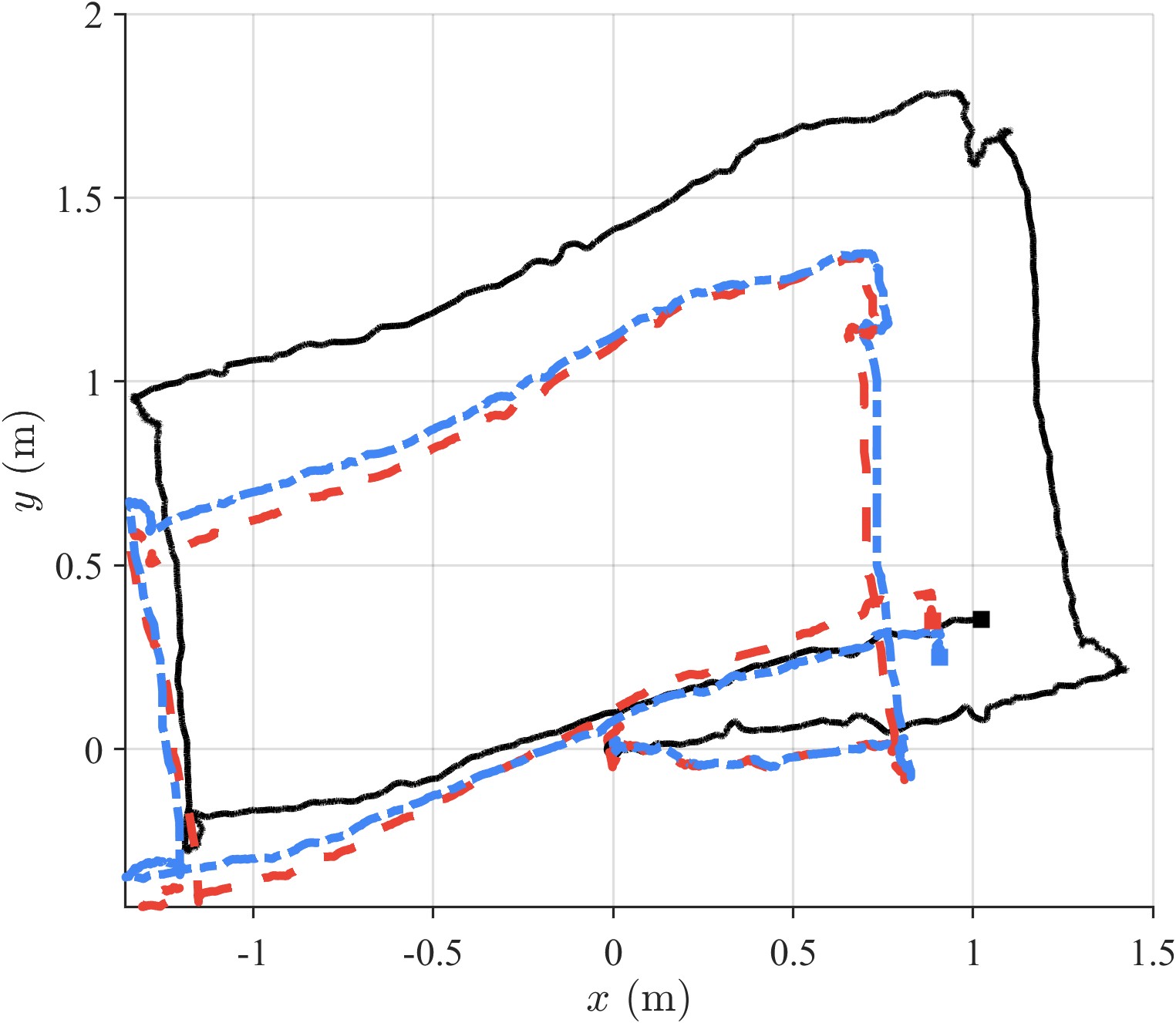}
    }
    \caption{ Position estimates obtained with IEKF and $\mathrm{SO}(3)$-EKF on real world public datasets with different robots: 
    a) ANYmal D \cite{grantour} ,  b) Unitree A1 \cite{yang2022}, c) ANYmal B \cite{camurri2017}, d) Unitree Go1 \cite{legkilo}, e) Unitree Go1 \cite{yang2023} 
      and f) Unitree Go2.}
    \label{fig:result_real_world_scenario}
\end{figure*}
In the final experiment, we explore one of the main features of the library: the ability to easily compare different filters on the same real-world dataset.
We extracted, from each one of the datasets described in Table \ref{tab:quadruped_datasets}, the proprioceptive data in the format defined in Table \ref{tab:proprioceptive_data}. 
We also included a new dataset collected with the Unitree Go2 robot. Once with the data, we ran both the $\mathrm{SO}(3)$-EKF and the IEKF with 
two consecutive update steps: measured base velocity and comparison with forward kinematics. The final config file was almost the same for each dataset, differing only in the path to the dataset and the URDF file of the robot. 
The results are shown in Fig. \ref{fig:result_real_world_scenario}. In most of the scenarios, it can be observed that 
IEKF outperformed $\mathrm{SO}(3)$-EKF. This becomes especially evident in Fig. ~\ref{fig:grantour}, \ref{fig:cerberus_1} and 
\ref{fig:legkilo}, where the $\mathrm{SO}(3)$-EKF estimated final position is far from the reference. 
This is one more consequence of the false observability issue, which causes the filter to become overconfident in its current state.
The final output of the estimation is a TUM \cite{sturm} trajectory file, which can be compared in terms of relative and absolute position error 
with the reference trajectory using evo~\cite{evo} or any similar package.
The parameters used in all estimations are given as follows:
\begin{equation}
    \label{eq:simulation_noise}
\begin{aligned}
\mathbf{P}_{0}
=
\mathbf{D}\!\left(
\left(\frac{\pi}{20}\right)^2 \mathbf{I}_3,\,
10^{-2}\mathbf{I}_3,\,
10^{-2}\mathbf{I}_3
\right), \\
\mathbf{w}_{\ell,i}
\sim
\mathcal{N}(\mathbf{0}_{3,1}, \mathbf{Q}_{\ell}),
\qquad
\ell \in \{a,g,b_a,b_g,v,f,k\}, \\
\mathbf{Q}_{a}
= \mathbf{D}(1.0,1.0,2.25),
\mathbf{Q}_{g}
= 2.5\times10^{-3}\mathbf{I}_3, \\
\mathbf{Q}_{b_a}
= 1.6\times10^{-7}\mathbf{I}_3,
\mathbf{Q}_{b_g}
= 4\times10^{-8}\mathbf{I}_3, \\
\mathbf{Q}_{v}
= \mathbf{D}(0.280,0.009,0.018), 
\mathbf{Q}_{f}
= \mathbf{Q}_{k}
= 10^{-4}\mathbf{I}_3,
\end{aligned}
\end{equation}
where $\mathbf{D} =: \mathrm{diag}(\cdot)$. 
%
%
\section{CONCLUSIONS}
We presented Chalito, a MATLAB and Python library offering a minimal, flexible
interface for implementing and benchmarking filtering algorithms for
quadruped state estimation in terms of accuracy and consistency,
demonstrated through three experiments on synthetic and real-world
data. Future work will add support for smoothing algorithms and
exteroceptive measurements, the latter potentially through a parallel C++ application
that processes raw exteroceptive data and exports objects implementing
the \texttt{MeasurementBase} interface, addressing the memory-efficiency
limits of exporting large exteroceptive datasets to a common format.






\bibliographystyle{IEEEtran}
\bibliography{main}


\end{document}